\documentclass{jpsj3}
\usepackage{bm}
\graphicspath{{figs/}}
\usepackage{amsfonts}
\usepackage{bm}
\usepackage{euscript}
\usepackage{amssymb}

\title{Bayesian image segmentations by Potts prior and loopy belief propagation}

\author{Kazuyuki Tanaka
\thanks{E-mail: kazu@smapip.is.tohoku.ac.jp}$^{1}$, 
\name{Shun Kataoka}$^{1}$, 
\name{Muneki Yasuda}$^{2}$, 
\name{Yuji Waizumi}$^{1}$ \\
and \name{Chiou-Ting Hsu}$^{3}$ 
}
\inst{
\address{$^{1}$Graduate School of Information Sciences,
Tohoku University, 6-3-09 Aramaki-aza-aoba,
Aoba-ku, Sendai 980-8579, Japan}\\
\address{$^{2}$Graduate School of Science and Engineering, 
Yamagata University, 
4-3-16 Jyounan, Yonezawa 992-8510, Japan}\\
\address{$^{3}$Department of Computer Science, 
National Tsing Hua University, 
No.101, Section 2, Kuang-Fu Road, Hsinchu, Taiwan 30013, R.O.C.}
} 

\abst{
This paper presents a Bayesian image segmentation model 
based on Potts prior and loopy belief propagation. 
The proposed Bayesian model involves several terms, 
including the pairwise interactions of Potts models, 
and the average vectors and covariant matrices of Gauss distributions 
in color image modeling. 
These terms are often referred to as hyperparameters 
in statistical machine learning theory. 
In order to determine these hyperparameters, 
we propose a new scheme for hyperparameter estimation 
based on conditional maximization 
of entropy in the Potts prior. The algorithm is given based on 
loopy belief propagation. In addition, 
we compare our conditional maximum entropy framework 
with the conventional maximum likelihood framework, 
and also clarify how the first order phase transitions 
in loopy belief propagations for Potts models 
influence our hyperparameter estimation procedures.
}

\kword{Bayesian statistics,
statistical-mechanical informatics,
maximum likelihood estimation,
probabilistic image processing, 
Markov random fields,
Potts model,
belief propagation,
advanced mean-field methods,
first-order phase transition}

\begin{document}

\maketitle

{\section{Introduction}} \label{sec:Introduction}

Bayesian image modeling based on Markov random fields (MRF) 
and loopy belief propagations (LBP) 
is one of the interesting research topics 
in statistical-mechanical 
informatics {\cite{Geman1990,Nishimori2001,
FreemanJonesPasztor2002,Willsky2002,KTanaka2002}}. 
Its advantages are two fold. 
First, Bayesian analysis provides useful statistical models 
for probabilistic information processing to treat massive 
and realistic datasets.
Second, statistical-mechanical informatics provides 
 powerful algorithms based on the advanced mean field methods, 
including the LBP, 
which is equivalent to the Bethe approximation 
in statistical mechanics
{\cite{KTanaka2002,KabashimaSaad1998,OpperSaad2001,
YedidiaFreemanWeiss2005,Pelizzola2005,
MezardMontanari2009}}.

Because MRF's usually include some hyperparameters 
which correspond to the temperature and interactions in
classical spin systems, one can determine these hyperparameters 
by maximizing marginal likelihoods in Bayesian modeling. 
The marginal likelihoods are constructed from probabilities 
of observed data with given hyperparameters 
and are expressed by free energies 
of prior and posterior probabilities. 
Practical algorithms can often be constructed 
based on the expectation-maximization (EM) 
algorithm{\cite{DempsterLairdRubinRoyal1977}}. 
From the statistical-mechanical stand-point, 
EM algorithms used in Bayesian image analysis have
been investigated by applying LBP 
to some classical spin systems{\cite{TanakaTitterington2007,
KataokaYasudaTanakaTitterington2012}}.
We have to mention that, in the EM algorithm, 
the differentiability of marginal likelihood 
with respect to hyperparameters is very important.
The classical spin systems 
in Refs.{\cite{TanakaTitterington2007,
KataokaYasudaTanakaTitterington2012}} have 
only second order phase transitions 
and the marginal likelihoods are always differentiable 
with respect to hyperparameters.

Image segmentation, 
as one of the primary but challenging topics in image processing, 
corresponds to the labeling of pixels 
in term of the three chromatic values 
at each pixel in the observed image.
Because image segmentation is usually defined 
on a finite square lattice of pixels, 
the MRF's can be formulated as having a high probability 
when the number of neighbouring pairs of pixels 
with the same labeling state is large{\cite{KatoZerubia2011}}. 
Such MRF modeling can be realized 
by considering ferromagnetic Potts models 
on the square lattice in the statistical mechanics.
The state at each pixel 
corresponds to the label in clustering the observed data. 
Bayesian modeling for image segmentations 
typically provides a posterior probabilistic model 
of labeling when a natural image is given.
It is often reduced to 
a $q$-state Potts model ($q=2,3,4,{\cdots}$) 
with spatially non-uniform external fields
and uniform nearest-neighbour interactions.

Various useful probabilistic inference algorithms 
for image segmentations 
have been proposed{\cite{LakshmananDerin1989,
Zhang1992,ZhangModestinoLangan1994,
DEliaPoggiScarpa2003,ChengJiaoSchuurmansWang-ICML2005,
ChenTanakaHoriguchi2005,McGroryTitteringtonReevesPettitt2009,
MiyoshiOkada2011,HasegawaOkadaMiyoshi2011,
GimenezFreryFlesia-IGARSS2013}} 
by means of the maximum likelihood framework for MRF's.
Particularly, inference algorithms in 
Refs.{\cite{Zhang1992,ChengJiaoSchuurmansWang-ICML2005,
ChenTanakaHoriguchi2005,McGroryTitteringtonReevesPettitt2009,
MiyoshiOkada2011,HasegawaOkadaMiyoshi2011,
GimenezFreryFlesia-IGARSS2013}} are based on 
advanced mean field methods, including the LBP; and
MRF's for image segmentations 
are using $q$-state Potts models 
as prior probabilities.
Carlucci and Inoue
adopted $q$-state Potts models 
with infinite-range interactions 
as prior probability 
distributions, and they investigated
statistical performance in Bayesian image modeling 
by using the replica 
method in the spin glass theory{\cite{CarlucciInoue1999}}.
As shown in Fig.{\ref{Figure01}},
it is known that, for $q$-state Potts model with $q{\ge}3$, 
the approximate free energies 
of the advanced mean field methods are continuous functions 
but have non-differentiable points with respect to 
the temperature{\cite{NishimoriOrtiz2011}}. 
Such singularities are often referred 
to as the first order phase transitions 
in the statistical mechanics.
Applications of LBP often 
leads to phase transitions 
for systems that include cycles in their graphical representations,
even if they are finite-size systems{\cite{TanakaKataokaYasuda2005}}. 
In Bayesian image restoration, 
the approximate marginal likelihood in LBP 
for three-state Potts prior has been computed 
for some artificial images and 
the above singularities have been shown to appear 
in the approximate marginal 
likelihood{\cite{TanakaTitterington2005}}.
Recently, an efficient iterative inference algorithm has been proposed
to realize the hyperparameter estimation 
in the standpoint of a conditional maximization 
of entropy for Bayesian image restoration 
by means of generalized sparse MRF prior 
and LBP{\cite{TanakaYasudaTitterington2012}. 
The scheme works well for prior probability 
with the first order phase transition. In addition, this scheme is equivalent 
to the EM algorithm for maximization of marginal likelihood 
when the differentiate of marginal likelihood with respect to hyperparameters 
is always a continuous function, and the prior probability has 
the second order phase transitions or no phase transitions. 
\begin{figure}
\begin{center}
\includegraphics[height=8.0cm, bb = 0 0 858 612]{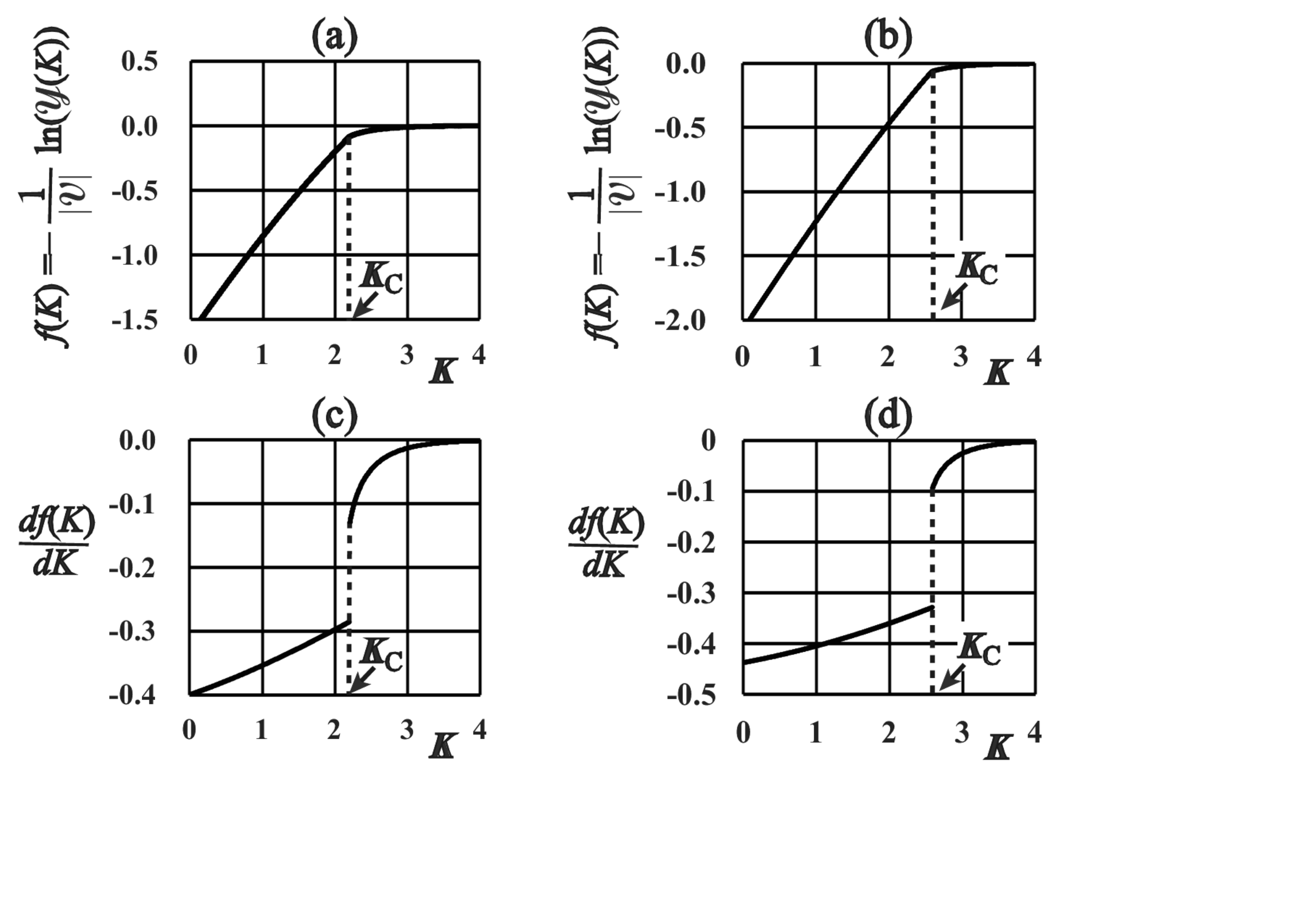}
\end{center}
\caption{
Free energy 
$f(K)
=-{\frac{1}{|{\cal{V}}|}}
{\ln}{\cal{Y}}(K)$ and the differentiate ${\frac{df(K)}{dK}}$ 
for various values of the inverse temperature $K(>0)$. 
They are obtained by applying the loopy belief propagation 
to $q$-state Potts models (\ref{Prior-ML}) 
on the square lattice 
with periodic boundary conditions along $x$- and $y$-directions.
(a) $f(K)$ for $q=5$.
(b) $f(K)$ for $q=8$.
(c) ${\frac{df(K)}{dK}}$ for $q=5$.
(b) ${\frac{df(K)}{dK}}$ for $q=8$.
Here 
${\cal{Y}}(K)$ 
is the partition function of the $q$-state Potts model 
in eq.(\ref{Prior-ML}), 
${\cal{V}}\equiv\{1,2,{\cdots},|{\cal{V}}|\}$ is the set of all the pixels and 
${\cal{E}}$ is the set of all the nearest neighbour pairs of pixels 
on the square lattice.
The first order transition points $K_{\rm{C}}$
of the Potts model 
in the loopy belief propagation are 
$2.1972$ and $2.5871$ for $q=5$ and $q=8$, respectively.
}
\label{Figure01}
\end{figure}

In the present paper, we will explain, for Bayesian image segmentation,  
how the first order phase transitions 
in LBP's for $q$-state Potts models influence 
EM algorithms in the maximum likelihood framework 
and how the inference algorithm 
in Ref.{\cite{TanakaYasudaTitterington2012}} 
works from the standpoint of statistical-mechanical informatics. 
In {\S}2, we construct 
a Potts prior probability distribution 
for Bayesian image segmentation modeling
from the standpoint of 
the constrained maximization of entropy.
In {\S}3, 
we propose a novel inference scheme, 
which is based on a conditional maximum likelihood framework, 
for estimating hyperparameters 
from an observed natural color image 
in terms of our Potts prior distribution and the LBP. 
In {\S}4, we survey the inference procedure 
for estimating hyperparameters 
using the conventional maximum likelihood framework
and give numerical experiments in the frameworks with the LBP.
We will also clarify how the first order phase transition 
appears in the conventional scheme with the LBP.
In {\S}5, we give some concluding remarks.

\vspace{5.0mm}

{\section{Potts Prior 
for Probabilistic Image 
Segmentation}} \label{sec: PottsPrior}

We consider an image 
as defined on a set of pixels 
arranged on a square grid graph $({\cal{V}},{\cal{E}})$, 
where ${\cal{V}}$ is the set of all the pixels and is defined 
by ${\cal{V}}{\equiv}\{i|i=1,2,{\cdots},|{\cal{V}}|\}$. 
There is a link $\{i,j\}$ 
between every nearest-neighbour pair 
of pixels $i$ and $j$, 
and ${\cal{E}}$ denotes the set of 
all the nearest-neighbour pairs 
of pixels $\{i,j\}$. 
The total numbers of elements 
in the sets ${\cal{V}}$ and ${\cal{E}}$ 
are denoted by $|{\cal{V}}|$ and $|{\cal{E}}|$, respectively.
The goal of image segmentation is to classify the pixels into  
several regions.
Each pixel will be assigned one 
of the integers ${\cal{Q}}{\equiv}\{0,1,2,{\cdots},q-1\}$ 
as its region label. 
In the present section, 
we give the prior probability distribution 
of labeled configurations  
on the square grid graph $({\cal{V}},{\cal{G}})$. 

The label at each pixel $i$ is regarded as a random variable, 
denoted by $A_{i}$.
Then the random field of labels is represented 
by ${\bm{A}}{\equiv}(A_{1},A_{2},{\cdots},A_{|{\cal{V}}|})^{\rm{T}}$,
and every labeled configuration is denoted 
by ${\bm{a}}=(a_{1},a_{2},{\cdots},a_{|{\cal{V}}|})^{\rm{T}}$. 
The prior probability 
of a labeled configuration ${\bm{a}}$ 
is assumed to be specified by a constant $u$ as
\begin{eqnarray}
& &{\hspace{-1.0cm}}
{\Pr}\{{\bm{A}}={\bm{a}}|u\}
= {\arg}{\max_{{\cal{P}}({\bm{a}})}}{\Big \{}
-{\sum_{\bm{z}}}{\cal{P}}({\bm{z}})
{\ln}{\cal{P}}({\bm{z}})
{\Big |}
\nonumber\\
& &
{\frac{1}{|{\cal{E}}|}}
{\sum_{\{i,j\}{\in}{\cal{E}}}}
{\sum_{{\bm{z}}}}
{\big (}1-{\delta}_{z_{i},z_{j}}{\big )}
{\cal{P}}({\bm{z}})=u,
{\sum_{\bm{z}}}
{\cal{P}}({\bm{z}})=1
{\Big \}},
\label{Prior-EntropyMax}
\end{eqnarray}
where 
${\bm{z}}=(z_{1},z_{2},{\cdots},z_{|{\cal{V}}|})^{\rm{T}}$ 
and 
${\displaystyle{{\sum_{\bm{z}}}{\equiv}
{\sum_{z_{1}{\in}{\cal{Q}}}}{\sum_{z_{2}{\in}{\cal{Q}}}}
{\cdots}{\sum_{z_{|{\cal{V}}|}{\in}{\cal{Q}}}}}}$. 
By introducing the Lagrange multipliers for the constraints, 
we reduce the prior probability 
${\Pr}\{{\bm{A}}={\bm{a}}|u\}$ to 
\begin{eqnarray}
{\Pr}\{{\bm{A}}={\bm{a}}|u\}
={\frac{1}{{\cal{Z}}(u)}}
{\prod_{\{i,j\}{\in}{\cal{E}}}}
{\exp}{\Big (}{\frac{1}{2}}{\alpha}(u)
{\delta}_{a_{i},a_{j}}{\Big )},
\label{Prior}
\end{eqnarray}
where ${\cal{Z}}(u)$ is a normalization constant. 
The interaction ${\alpha}(u)$ 
is a function of $u$ 
and should be determined to satisfy 
the following constraint condition 
\begin{eqnarray}
{\frac{1}{|{\cal{E}}|}}
{\sum_{\{i,j\}{\in}{\cal{E}}}}
{\sum_{\bm{z}}}{\big (}1-{\delta}_{z_{i},z_{j}}{\big )}
{\Pr}\{{\bm{A}}={\bm{z}}|u\}=u.
\label{PriorConstraint}
\end{eqnarray}
In order to calculate the estimate 
of the hyperparameter 
${\alpha}(u)$, 
we have to solve the following equation: 
\begin{eqnarray}
& &
{\frac{1}{|{\cal{E}}|}}{\sum_{\{i,j\}{\in}{\cal{E}}}}
{\sum_{{\zeta}{\in}{\cal{Q}}}}
{\sum_{{\zeta}'{\in}{\cal{Q}}}}
{\big (}1-{\delta}_{{\zeta},{\zeta}'}{\big )}
{\Pr}\{A_{i}={\zeta},A_{j}={\zeta}'|u\}
=u,
\label{alpha-determination}
\end{eqnarray}
where
\begin{eqnarray}
{\Pr}\{A_{i}=a_{i},A_{j}=a_{j}|u\}
\equiv 
{\sum_{\bm{z}}}{\delta}_{z_{i},a_{i}}{\delta}_{z_{j},a_{j}}
{\Pr}\{{\bm{A}}={\bm{z}}|u\}.
\label{PriorMarginalProbability-Edge}
\end{eqnarray}

In the above mathematical framework, as shown in
the deterministic equation 
(\ref{alpha-determination}) 
together with eqs.(\ref{Prior})
and (\ref{PriorMarginalProbability-Edge}),
computation of the two terms
${\sum_{\{i,j\}{\in}{\cal{E}}}}
{\sum_{\bm{z}}}{\big (}1-{\delta}_{z_{i},z_{j}}{\big )}
{\Pr}\{{\bm{A}}={\bm{z}}|u\}$ 
and ${\cal{Z}}(u)$ is critical to ${\alpha}(u)$.
In LBP{\cite{YedidiaFreemanWeiss2005,Pelizzola2005,
MezardMontanari2009,TanakaYasudaTitterington2012}}, 
the marginal prior 
probability distributions 
${\Pr}\{A_{i}=a_{i},A_{j}=a_{j}|u\}$ 
in eq.(\ref{PriorMarginalProbability-Edge}) 
and 
${\Pr}\{A_{i}=a_{i}|u\}
\equiv {\sum_{{\bm{z}}}}
{\delta}_{z_{i},a_{i}}
{\Pr}\{{\bm{A}}={\bm{z}}|u\}$ 
can be approximately reduced to 
\begin{eqnarray}
{\Pr}\{A_{i}=a_{i},A_{j}=a_{j}|u\}
& \simeq &
{\frac{1}{{\cal{Z}}_{\{i,j\}}(u)}}
{\Big (}
{\prod_{k{\in}{\partial}i{\backslash}\{j\}}}
{\lambda}_{k{\to}i}(a_{i})
{\Big )}
{\exp}{\Big (}{\frac{1}{2}}{\alpha}(u)
{\delta}_{a_{i},a_{j}}{\Big )}
\nonumber\\
& &{\hspace{1.50cm}}{\times}
{\Big (}
{\prod_{k{\in}{\partial}j{\backslash}\{i\}}}
{\lambda}_{k{\to}j}(a_{j})
{\Big )}
~(\{i,j\}{\in}{\cal{E}}),
\label{LBP-PriorEdge}
\end{eqnarray}
\begin{eqnarray}
{\Pr}\{A_{i}=a_{i}|u\}
\simeq
{\frac{1}{{\cal{Z}}_{i}(u)}}
{\prod_{k{\in}{\partial}i}}
{\lambda}_{k{\to}i}(a_{i})
{\hspace{2.0mm}}(i{\in}{\cal{V}}),
\label{LBP-PriorNode}
\end{eqnarray}
where ${\partial}i$ denotes the set of neighbouring pixels of pixel $i$.
The quantities
${\cal{Z}}_{\{i,j\}}(u)$ 
and ${\cal{Z}}_{i}(u)$ 
in eqs.(\ref{LBP-PriorEdge})
and (\ref{LBP-PriorNode})
correspond to normalization constants 
of approximate representations 
of marginal probabilities in LBP. 
Here $\{{\lambda}_{j{\to}i}({\xi})|
i{\in}{\cal{V}},~j{\in}{\partial}i,~{\xi}{\in}{\cal{Q}}\}$ 
are messages 
in the LBP{\cite{TanakaYasudaTitterington2012}} 
for the prior probability 
${\Pr}\{{\bm{A}}={\bm{a}}|u\}$ in eq.(\ref{Prior}), 
and the free energy $f(u)$ per pixel in the Potts prior (\ref{Prior}) 
is also approximately expressed as
\begin{eqnarray}
f(u)
& \equiv &
-{\frac{1}{|{\cal{V}}|}}
{\ln}{\cal{Z}}(u)
\nonumber\\
&{\simeq}&
{\frac{1}{|{\cal{V}}|}}
{\Big (}
-{\sum_{\{i,j\}{\in}{\cal{E}}}}{\ln}{\cal{Z}}_{\{i,j\}}(u)
-{\sum_{i{\in}{\cal{V}}}}(1-|{\partial}i|){\ln}{\cal{Z}}_{i}(u)
{\Big )}.
\label{LBPPriorPartitionFunction}
\end{eqnarray}
The messages 
${\lambda}_{j{\to}i}({\xi})$ 
(${\xi}{\in}{\cal{Q}}$, 
$j{\in}{\partial}i$, $i{\in}{\cal{V}}$) 
are determined so as to satisfy 
the following simultaneous equations:
\begin{eqnarray}
& &{\hspace{-0.50cm}}
{\lambda}_{j{\to}i}({\xi})
=
{\frac{
{\displaystyle{{\sum_{{\zeta}{\in}{\cal{Q}}}}}}
{\exp}{\Big (}{\frac{1}{2}}{\alpha}(u){\delta}_{{\xi},{\zeta}}{\Big )}
{\displaystyle{
{\prod_{k{\in}{\partial}j{\backslash}\{i\}}}
}}
{\lambda}_{k{\to}j}({\zeta})
}
{
{\displaystyle{{\sum_{{\zeta}{\in}{\cal{Q}}}}}}
{\displaystyle{{\sum_{{\zeta}'{\in}{\cal{Q}}}}}}
{\exp}{\Big (}{\frac{1}{2}}{\alpha}(u){\delta}_{{\zeta}',{\zeta}}{\Big )}
{\displaystyle{
{\prod_{k{\in}{\partial}j{\backslash}\{i\}}}
}}
{\lambda}_{k{\to}j}({\zeta})
}
}
\nonumber\\
& &{\hspace{3.0cm}}
(i{\in}{\cal{V}},
~j{\in}{\partial}i,
~{\xi}{\in}{\cal{Q}}).
\label{LBPMessagePassingRule}
\end{eqnarray}

In Fig.{\ref{Figure02}}, we show the curves of ${\alpha}(u)$ 
and $f(u)$ along various values of $u$ 
for $q=5$ and $q=8$.
For each fixed value of $u$, ${\alpha}(u)$ is determined so as to satisfy 
the constraint condition (\ref{alpha-determination}).
The left-hand side 
of the constraint condition (\ref{alpha-determination}) is computed 
by using eq.(\ref{LBP-PriorEdge}) together 
with eq.(\ref{LBPMessagePassingRule}) in LBP.
\begin{figure}
\begin{center}
\includegraphics[height=8.0cm, bb = 0 0 858 612]{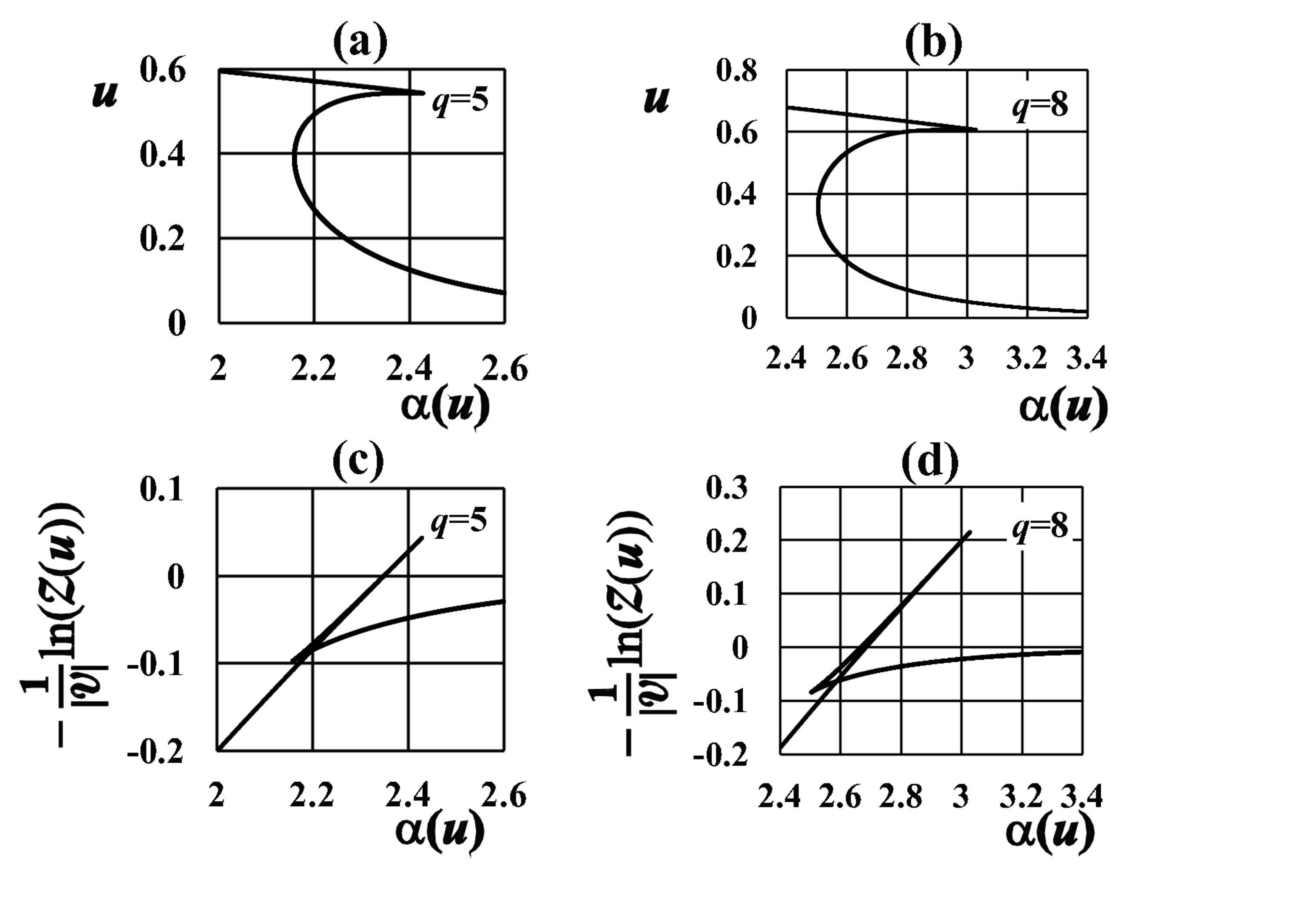}
\end{center}
\caption{
${\alpha}(u)$ and $f(u)
=-{\frac{1}{|{\cal{V}}|}}
{\ln}{\cal{Z}}(u)$ for various values of $u$ 
obtained by using the loopy belief propagation of Potts models 
for the cases of $q=5$ and $q=8$.
}
\label{Figure02}
\end{figure}

\vspace{5.0mm}

{\section{Segmentation Algorithm 
for Potts Posterior 
and Loopy Belief 
Propagation}} \label{sec: ImageSegmentationModeling}

In this section, 
we provide a posterior probability 
and a hyperparameter estimation 
scheme in terms of 
the Potts prior constructed
in the previous section.
We combine the conditional maximization of entropy 
with Bayesian modeling to derive simultaneous deterministic equations 
for estimating hyperparameters from the given data.

The intensities of 
red, green, and blue channels at pixel $i$ 
in the observed image 
are regarded as random variables 
denoted by 
$D_{i}^{\rm{R}}$, $D_{i}^{\rm{G}}$ 
and $D_{i}^{\rm{B}}$, respectively.
The random fields of red, green and blue intensities 
in the observed color image are then represented
by the $3|{\cal{V}}|$-dimensional vector 
${\bm{D}}{\equiv}({\bm{D}}_{1},{\bm{D}}_{2},
{\cdots},{\bm{D}}_{|{\cal{V}}|})^{\rm{T}}$, 
where ${\bm{D}}_{i}
\equiv (D_{i}^{\rm{R}},D_{i}^{\rm{G}},
D_{i}^{\rm{B}})^{\rm{T}}$.
The actual color image is denoted by 
${\bm{d}}=({\bm{d}}_{1},{\bm{d}}_{2},
{\cdots},{\bm{d}}_{|{\cal{V}}|})^{\rm{T}}$, 
where ${\bm{d}}_{i}
=(d_{i}^{\rm{R}},d_{i}^{\rm{G}},
d_{i}^{\rm{B}})^{\rm{T}}$.
The random variables $D_{i}^{\rm{R}}$, $D_{i}^{\rm{G}}$ 
and $D_{i}^{\rm{B}}$ at each pixel $i$ can take 
any real numbers 
in the interval $(-{\infty},+{\infty})$.
The generative process of natural color images ${\bm{d}}$ 
is assumed to be the following conditional probability:
\begin{eqnarray}
& &{\hspace{-1.0cm}}
{\Pr}\{{\bm{D}}={\bm{d}}|
{\bm{A}}={\bm{a}},{\bm{\Theta}}\}
= 
{\prod_{i{\in}{\cal{V}}}}
g({\bm{d}}_{i}|a_{i},{\bm{\Theta}}),
\label{DegradationProcess}
\end{eqnarray}
where 
\begin{eqnarray}
{\bm{\Theta}}
\equiv
{\Bigg \{}
{\bm{m}}({\xi}) = 
 \left(
   \begin{array}{@{\,}ccc@{\,}}
    m_{\rm{R}}({\xi}) \\
    m_{\rm{G}}({\xi}) \\
    m_{\rm{B}}({\xi}) 
  \end{array}
 \right)
,
{\bm{C}}({\xi}) = 
 \left(
   \begin{array}{@{\,}ccc@{\,}}
    C_{\rm{RR}}({\xi}) 
          & C_{\rm{GR}}({\xi}) 
          & C_{\rm{BR}}({\xi}) \\
    C_{\rm{RG}}({\xi}) 
          & C_{\rm{GG}}({\xi}) 
          & C_{\rm{BG}}({\xi}) \\
    C_{\rm{RB}}({\xi}) 
          & C_{\rm{GB}}({\xi}) 
          & C_{\rm{BB}}({\xi})
  \end{array}
 \right)
{\Bigg |}{\xi}{\in}{\cal{Q}}
{\Bigg \}}
\end{eqnarray}
and 
\begin{eqnarray}
& &{\hspace{-1.0cm}}
g({\bm{d}}_{i}|a_{i},{\bm{\Theta}})
\equiv 
{\sqrt{{\frac{1}{{\rm{det}}(2{\pi}{\bm{C}}(a_{i}))}}}}
{\exp}{\Big (}
-{\frac{1}{2}}
({\bm{d}}_{i}-{\bm{m}}(a_{i}))^{\rm{T}}
{\bm{C}}^{-1}(a_{i})
({\bm{d}}_{i}-{\bm{m}}(a_{i}))
{\Big )}.
\label{2DGaussian}
\end{eqnarray}

By substituting 
eqs.(\ref{Prior}) and 
(\ref{DegradationProcess}) 
into the Bayes formula, 
we derive the posterior probability distribution as follows:
\begin{eqnarray}
{\Pr}\{{\bm{A}}={\bm{a}}|
{\bm{D}}={\bm{d}},u,{\bm{\Theta}}\}
={\frac{1}{{\cal{Z}}({\bm{d}},u,{\bm{\Theta}})}}
{\Big (}{\prod_{i{\in}{\cal{V}}}}g({\bm{d}}_{i}|a_{i},{\bm{\Theta}}){\Big )}
{\Big (}{\prod_{\{i,j\}{\in}{\cal{E}}}}
{\exp}{\big (}{\frac{1}{2}}{\cal{K}}{\delta}_{a_{i},a_{j}}{\big )}{\Big )},
\label{Posterior}
\end{eqnarray}
where
\begin{eqnarray}
{\cal{Z}}({\bm{d}},u,{\bm{\Theta}})
\equiv 
{\sum_{{\bm{z}}}}
{\Big (}{\prod_{i{\in}{\cal{V}}}}g({\bm{d}}_{i}|a_{i},{\bm{\Theta}}){\Big )}
{\Big (}{\prod_{\{i,j\}{\in}{\cal{E}}}}
{\exp}{\big (}{\frac{1}{2}}{\cal{K}}{\delta}_{a_{i},a_{j}}{\big )}{\Big )}.
\label{Posterior-PartitionFunction}
\end{eqnarray}
Another way of defining the posterior probability 
of a labeling ${\bm{a}}$ can be introduced 
through the following definition:
\begin{eqnarray}
& &{\hspace{-1.50cm}}
{\Pr}\{{\bm{A}}={\bm{a}}|{\bm{D}}={\bm{d}},u,{\bm{\Theta}}\}
= 
{\frac{
{\Big (}{\displaystyle{{\prod_{i{\in}V}}}}
{\sqrt{{\frac{1}{{\rm{det}}(2{\pi}{\bm{C}}(a_{i}))}}}}{\Big )}
{\widehat{{\cal{P}}}}({\bm{a}})
}{
{\displaystyle{{\sum_{{\bm{z}}}}}}
{\Big (}{\displaystyle{{\prod_{i{\in}V}}}}
{\sqrt{{\frac{1}{{\rm{det}}(2{\pi}{\bm{C}}(z_{i}))}}}}{\Big )}
{\widehat{{\cal{P}}}}({\bm{z}})
}}
\label{Posterior-EntropyMaxB}
\end{eqnarray}
\begin{eqnarray}
& &{\hspace{-1.50cm}}
{\widehat{{\cal{P}}}}({\bm{a}})
 = 
{\arg}{\max_{{\cal{P}}({\bm{a}})}}{\Big \{}
-{\sum_{\bm{z}}}{\cal{P}}({\bm{z}})
{\ln}{\cal{P}}({\bm{z}})
{\Big |}
{\sum_{\bm{z}}}{\cal{P}}({\bm{z}})=1,
{\sum_{\{i,j\}{\in}{\cal{E}}}}
{\sum_{\bm{z}}}
{\big (}1-{\delta}_{z_{i},z_{j}}{\big )}
{\cal{P}}({\bm{z}})=u,
\nonumber\\
& &{\hspace{-0.50cm}}
{\sum_{i{\in}{\cal{V}}}}
{\bm{d}}_{i}
{\sum_{\bm{z}}}
{\delta}_{z_{i},{\xi}}{\cal{P}}({\bm{z}})
=
{\sum_{i{\in}{\cal{V}}}}
{\bm{m}}({\xi})
{\sum_{\bm{z}}}
{\delta}_{z_{i},{\xi}}{\cal{P}}({\bm{z}})
~({\xi}{\in}{\cal{Q}}),
\nonumber\\
& &{\hspace{-0.50cm}}
{\sum_{i{\in}{\cal{V}}}}
{\bm{d}}_{i}{\bm{d}}_{i}^{\rm{T}}
{\displaystyle{{\sum_{\bm{z}}}}}
{\delta}_{z_{i},{\xi}}
{\cal{P}}({\bm{z}})
=
{\sum_{i{\in}{\cal{V}}}}
{\big (}{\bm{C}}({\xi})
+{\bm{m}}({\xi}){\bm{m}}^{\rm{T}}({\xi}){\big )}
{\displaystyle{{\sum_{\bm{z}}}}}
{\delta}_{z_{i},{\xi}}
{\cal{P}}({\bm{z}})
~({\xi}{\in}{\cal{Q}})
{\Big \}}.
\label{Posterior-EntropyMax}
\end{eqnarray}
By introducing Lagrange multipliers 
${\Lambda}_{0}$,
${\Lambda}_{1}$,
${\bm{\Lambda}}_{2}({\xi})=
 \left(
   \begin{array}{@{\,}ccc@{\,}}
    {\Lambda}_{2}^{\rm{R}}({\xi}) \\
    {\Lambda}_{2}^{\rm{G}}({\xi}) \\
    {\Lambda}_{2}^{\rm{B}}({\xi}) 
  \end{array}
 \right)$ (${\xi}{\in}{\cal{Q}}$) and 
${\bm{\Lambda}}_{3}({\xi})=
 \left(
   \begin{array}{@{\,}ccc@{\,}}
\footnotesize
    {\Lambda}_{3}^{\rm{RR}}({\xi}) 
          & {\Lambda}_{3}^{\rm{GR}}({\xi}) 
          & {\Lambda}_{3}^{\rm{BR}}({\xi}) \\
    {\Lambda}_{3}^{\rm{RG}}({\xi}) 
          & {\Lambda}_{3}^{\rm{GG}}({\xi}) 
          & {\Lambda}_{3}^{\rm{BG}}({\xi}) \\
    {\Lambda}_{3}^{\rm{RB}}({\xi}) 
          & {\Lambda}_{3}^{\rm{GB}}({\xi}) 
          & {\Lambda}_{3}^{\rm{BB}}({\xi})
  \end{array}
 \right)$  (${\xi}{\in}{\cal{Q}}$) 
for the constraints and by considering the extremum condition 
with respect to ${\cal{P}}({\bm{a}})$,
the right-hand side of eq.({\ref{Posterior-EntropyMax}}) 
is reduced to the following expression: 
\begin{eqnarray}
{\cal{P}}({\bm{a}}) & \propto &
{\Big (}
{\prod_{\{i,j\}{\in}{\cal{E}}}}
{\exp}{\big (}-{\Lambda}_{1}(1-{\delta}_{a_{i},a_{j}}){\big )}
{\Big )}
\nonumber\\
& &{\hspace{-0.50cm}}
{\times}{\Bigg (}
{\prod_{i{\in}{\cal{V}}}}
{\exp}{\Big (}
-{\bm{d}}_{i}^{\rm{T}}{\bm{{\Lambda}}}_{3}(a_{i}){\bm{d}}_{i}
-{\frac{1}{2}}{\bm{d}}_{i}^{\rm{T}}{\bm{{\Lambda}}}_{2}(a_{i})
-{\frac{1}{2}}{\bm{{\Lambda}}}_{2}^{\rm{T}}(a_{i}){\bm{d}}_{i}^{\rm{T}}
{\Big )}
\nonumber\\
& &{\hspace{-0.50cm}}
{\times}
{\exp}{\Big (}
{\bm{m}}^{\rm{T}}(a_{i}){\bm{{\Lambda}}}_{3}(a_{i}){\bm{m}}(a_{i})
+{\frac{1}{2}}{\bm{{\Lambda}}}_{2}^{\rm{T}}(a_{i}){\bm{m}}(a_{i})
+{\frac{1}{2}}{\bm{m}}^{\rm{T}}(a_{i}){\bm{{\Lambda}}}_{2}(a_{i})
{\Big )}
\nonumber\\
& &{\hspace{-0.50cm}}
{\times}
{\exp}{\Big (}
{\sum_{{\kappa}{\in}\{{\rm{R}},{\rm{G}},{\rm{B}}\}}}
{\sum_{{\kappa}'{\in}\{{\rm{R}},{\rm{G}},{\rm{B}}\}}}
C_{{\kappa},{\kappa}'}(a_{i}){\Lambda}_{3}^{{\kappa},{\kappa}'}(a_{i})
{\Big )}
{\Bigg )},
\label{Posterior-EntropyMax-B}
\end{eqnarray}
up to the normalization constant including ${\Lambda}_{0}$.
The Lagrange multipliers 
${\Lambda}_{1}$, 
${\bm{{\Lambda}}}_{3}({\xi})$  (${\xi}{\in}{\cal{Q}}$) 
and ${\bm{{\Lambda}}}_{2}({\xi})$ (${\xi}{\in}{\cal{Q}}$) 
are determined so as to satisfy the constraint conditions: 
\begin{eqnarray}
& &
{\frac{1}{|{\cal{E}}|}}
{\displaystyle{{\sum_{\{i,j\}{\in}{\cal{E}}}}}}
{\displaystyle{{\sum_{\bm{z}}}}}
{\big (}1-{\delta}_{z_{i},z_{j}}{\big )}
{\cal{P}}({\bm{z}})=u,
\label{Posterior-EntropyMax-ConstraintsA}
\\
& &
{\frac{
{\displaystyle{{\sum_{i{\in}{\cal{V}}}}}}
{\bm{d}}_{i}
{\displaystyle{{\sum_{\bm{z}}}}}
{\delta}_{z_{i},{\xi}}{\cal{P}}({\bm{z}})
}{
{\displaystyle{{\sum_{i{\in}{\cal{V}}}}}}
{\displaystyle{{\sum_{\bm{z}}}}}
{\delta}_{z_{i},{\xi}}{\cal{P}}({\bm{z}})
}}
={\bm{m}}({\xi})~({\xi}{\in}{\cal{Q}}),
\label{Posterior-EntropyMax-ConstraintsB}
\\
& &
{\frac{
{\displaystyle{{\sum_{i{\in}{\cal{V}}}}}}
{\big (}
{\bm{d}}_{i}-{\bm{m}}({\xi}){\big )}
{\big (}{\bm{d}}_{i}-{\bm{m}}({\xi}){\big )}^{\rm{T}}
{\displaystyle{{\sum_{\bm{z}}}}}
{\delta}_{z_{i},{\xi}}
{\cal{P}}({\bm{z}})
}{
{\displaystyle{{\sum_{i{\in}{\cal{V}}}}}}
{\displaystyle{{\sum_{\bm{z}}}}}
{\delta}_{z_{i},{\xi}}
{\cal{P}}({\bm{z}})
}}
={\bm{C}}({\xi})
~({\xi}{\in}{\cal{Q}}).
\label{Posterior-EntropyMax-ConstraintsC}
\end{eqnarray}
Moreover, in order to ensure 
eq.(\ref{Posterior-EntropyMax-B}) as an identity 
with respect to every label configuration of ${\bm{a}}$, 
we have to impose the following equalities:
\begin{eqnarray}
& &
{\Lambda}_{1}={\frac{1}{2}}{\alpha}(u),
~({\xi}{\in}{\cal{Q}}),
\label{Posterior-EntropyMax-SufficientCondistionsA}
\\
& &{\bm{\Lambda}}_{2}({\xi})=-{\bm{C}}^{-1}({\xi}){\bm{m}}({\xi})
~({\xi}{\in}{\cal{Q}}),
\label{Posterior-EntropyMax-SufficientCondistionsB}
\\
& &{{\bm{{\Lambda}}}_{3}}({\xi})={\frac{1}{2}}{\bm{C}}^{-1}({\xi})
~({\xi}{\in}{\cal{Q}}),
\label{Posterior-EntropyMax-SufficientCondistionsC}
\end{eqnarray}
as sufficient conditions 
for eq.(\ref{Posterior-EntropyMaxB}) 
with respect to the right-hand sides of equations 
(\ref{Posterior}) and (\ref{Posterior-EntropyMax-B}).
Because ${\bm{C}}({\xi})$ (${\xi}{\in}{\cal{Q}}$) are symmetric matrices, 
we can show that ${\displaystyle{{\sum_{{\kappa}{\in}\{{\rm{R}},{\rm{G}},{\rm{B}}\}}}}}
{\displaystyle{{\sum_{{\kappa}'{\in}\{{\rm{R}},{\rm{G}},{\rm{B}}\}}}}}
 C_{{\kappa},{\kappa}'}({\xi}){\Lambda}_{3}^{{\kappa},{\kappa}'}({\xi})
={\frac{3}{2}}$ (${\xi}{\in}{\cal{Q}}$) 
in eq.(\ref{Posterior-EntropyMax-B}) by using 
eq.(\ref{Posterior-EntropyMax-SufficientCondistionsC}).
By combining the above arguments 
(\ref{Posterior}), 
(\ref{Posterior-EntropyMax-B}),
(\ref{Posterior-EntropyMax-ConstraintsA})-(\ref{Posterior-EntropyMax-ConstraintsC}),
and 
(\ref{Posterior-EntropyMax-SufficientCondistionsA})-(\ref{Posterior-EntropyMax-SufficientCondistionsB}) 
with the ones in eq.(\ref{Prior}) and eq.(\ref{PriorConstraint}), 
the simultaneous deterministic equations 
of estimates ${\widehat{u}}({\bm{d}})$ and 
${\widehat{\bm{\Theta}}}({\bm{d}})=
\{{\widehat{\bm{m}}}({\xi},{\bm{d}}),
{\widehat{\bm{C}}}({\xi},{\bm{d}})|{\xi}{\in}{\cal{Q}}\}$
of $u$ and ${\bm{\Theta}}$ should then be reduced 
to the following constraints:
\begin{eqnarray}
& &{\hspace{-1.0cm}}
{\frac{1}{|{\cal{E}}|}}
{\sum_{\{i,j\}{\in}{\cal{E}}}}
{\sum_{{\zeta}{\in}{\cal{Q}}}}
{\sum_{{\zeta}'{\in}{\cal{Q}}}}
{\big (}1-{\delta}_{{\zeta},{\zeta}'}{\big )}
{\Pr}\{A_{i}={\zeta},A_{j}={\zeta}'|
{\bm{D}}={\bm{d}},{\widehat{u}}({\bm{d}}),{\widehat{\bm{\Theta}}}({\bm{d}})\}
\nonumber\\
& = &
{\frac{1}{|{\cal{E}}|}}
{\sum_{\{i,j\}{\in}{\cal{E}}}}
{\sum_{{\zeta}{\in}{\cal{Q}}}}
{\sum_{{\zeta}'{\in}{\cal{Q}}}}
{\big (}1-{\delta}_{{\zeta},{\zeta}'}{\big )}
{\Pr}\{A_{i}={\zeta},A_{j}={\zeta}'|
{\widehat{u}}({\bm{d}})\},
\label{Determination-u-posterior}
\end{eqnarray}
\begin{eqnarray}
& &{\hspace{-1.0cm}}
{\frac{
{\displaystyle{{\sum_{i{\in}{\cal{V}}}}}}{\bm{d}}_{i}
{\Pr}\{A_{i}={\xi}|
{\bm{D}}={\bm{d}},{\widehat{u}}({\bm{d}}),{\widehat{\bm{\Theta}}}({\bm{d}})\}
}
{
{\displaystyle{{\sum_{i{\in}{\cal{V}}}}}}
{\Pr}\{A_{i}={\xi}|
{\bm{D}}={\bm{d}},{\widehat{u}}({\bm{d}}),{\widehat{\bm{\Theta}}}({\bm{d}})\}
}}
={\widehat{\bm{m}}}({\xi},{\bm{d}})
~({\xi}{\in}{\cal{Q}}),
\label{Determination-m-posterior}
\end{eqnarray}
\begin{eqnarray}
& &{\hspace{-1.0cm}}
{\frac{
{\displaystyle{{\sum_{i{\in}{\cal{V}}}}}}
{\big (}{\bm{d}}_{i}-{\widehat{\bm{m}}}({\xi},{\bm{d}}){\big )}
{\big (}{\bm{d}}_{i}-{\widehat{\bm{m}}}({\xi},{\bm{d}}){\big )}^{\rm{T}}
{\Pr}\{A_{i}={\xi}|
{\bm{D}}={\bm{d}},{\widehat{u}}({\bm{d}}),{\widehat{\bm{\Theta}}}({\bm{d}})\}
}
{
{\displaystyle{{\sum_{i{\in}{\cal{V}}}}}}
{\Pr}\{A_{i}={\xi}|
{\bm{D}}={\bm{d}},{\widehat{u}}({\bm{d}}),{\widehat{\bm{\Theta}}}({\bm{d}})\}
}}
\nonumber\\
& &{\hspace{7.0cm}}
={\widehat{\bm{C}}}({\xi},{\bm{d}})
~({\xi}{\in}{\cal{Q}}),
\label{Determination-C-posterior}
\end{eqnarray}
where
\begin{eqnarray}
{\Pr}\{A_{i}=a_{i},A_{j}=a_{j}|
{\bm{D}}={\bm{d}},u,{\bm{\Theta}}\}
\equiv 
{\sum_{\bm{z}}}{\delta}_{z_{i},a_{i}}{\delta}_{z_{j},a_{j}}
{\Pr}\{{\bm{A}}={\bm{z}}|
{\bm{D}}={\bm{d}},u,{\bm{\Theta}}\},
\end{eqnarray}
\begin{eqnarray}
{\Pr}\{A_{i}=a_{i}|
{\bm{D}}={\bm{d}},u,{\bm{\Theta}}\}
\equiv 
{\sum_{\bm{z}}}{\delta}_{z_{i},a_{i}}
{\Pr}\{{\bm{A}}={\bm{z}}
|{\bm{D}}={\bm{d}},u,{\bm{\Theta}}\}.
\end{eqnarray}
Given the estimates 
${\widehat{u}}$ and ${\widehat{\bm{\Theta}}}$,
the estimate of labeling 
${\widehat{\bm{a}}}({\bm{d}})
=({\widehat{a}}_{1}({\bm{d}}),
{\widehat{a}}_{2}({\bm{d}}),
{\cdots},
{\widehat{a}}_{|{\cal{V}}|}({\bm{d}}))^{\rm{T}}$ 
is determined by
\begin{eqnarray}
{\widehat{a}}_{i}({\bm{d}})
\equiv
{\max_{{\zeta}{\in}{\cal{Q}}}}
   {\Pr}\{A_{i}={\zeta}|
   {\bm{D}}={\bm{d}},
   {\widehat{u}}({\bm{d}}),{\widehat{\bm{\Theta}}}({\bm{d}})\}
   {\hspace{2.0mm}}(i{\in}{\cal{V}}).
\label{MPM}
\end{eqnarray}
The above method of producing 
the labeling is called 
maximum posterior marginal (MPM) estimation. 

In LBP, the marginal probability 
distributions 
${\Pr}\{A_{i}=a_{i},A_{j}=a_{j}
|{\bm{D}}={\bm{d}},u,{\bm{\Theta}}\}$ and 
${\Pr}\{A_{i}=a_{i}|{\bm{D}}={\bm{d}},u,{\bm{\Theta}}\}$ 
can be approximately reduced to 
\begin{eqnarray}
{\Pr}\{A_{i}=a_{i},A_{j}=a_{j}
|{\bm{D}}={\bm{d}},u,{\bm{\Theta}}\}
& \simeq & 
{\frac{1}{{\cal{Z}}_{\{i,j\}}({\bm{d}},u,{\bm{\Theta}})}}
{\Big (}
{\prod_{k{\in}
{\partial}i{\backslash}\{j\}}}
{\mu}_{k{\to}i}(a_{i})
{\Big )}
\nonumber\\
& &{\hspace{-6.0cm}}
{\times}
g({\bm{d}}_{i}|a_{i},{\bm{\Theta}})
{\exp}{\big (}{\frac{1}{2}}{\alpha}(u){\delta}_{a_{i},a_{j}}{\big )}
g({\bm{d}}_{j}|a_{j},{\bm{\Theta}})
{\Big (}
{\prod_{k{\in}{\partial}j{\backslash}\{i\}}}
{\mu}_{k{\to}j}(a_{j})
{\Big )}{\hspace{2.0mm}}(\{i,j\}{\in}{\cal{E}}),
\label{Determination-u-posteriorB}
\end{eqnarray}
\begin{eqnarray}
{\Pr}\{A_{i}=a_{i}
|{\bm{D}}={\bm{d}},u,{\bm{\Theta}}\}
& \simeq & 
{\frac{1}{{\cal{Z}}_{i}({\bm{d}},u,{\bm{\Theta}})}}
g({\bm{d}}_{i}|a_{i},{\bm{\Theta}})
{\Big (}
{\prod_{k{\in}{\partial}i}}
{\mu}_{k{\to}i}(a_{i}){\Big )}
{\hspace{2.0mm}}(i{\in}{\cal{V}}).
\label{Determination-m-posteriorB}
\end{eqnarray}
The quantities
${\cal{Z}}_{\{i,j\}}({\bm{d}},u,{\bm{\Theta}})$ and 
${\cal{Z}}_{i}({\bm{d}},u,{\bm{\Theta}})$ 
in eqs.(\ref{Determination-u-posteriorB})
and (\ref{Determination-m-posteriorB})
correspond to normalization constants 
of approximate representation 
to marginal probabilities in LBP. 
Here $\{{\mu}_{j{\to}i}({\xi})|
i{\in}{\cal{V}},~j{\in}{\partial}i,~{\xi}{\in}{\cal{Q}}\}$ 
are messages 
in the LBP{\cite{TanakaYasudaTitterington2012}} 
for the posterior probabilities 
${\Pr}\{{\bm{A}}={\bm{a}}|
{\bm{D}}={\bm{d}},u,{\bm{\Theta}}\}$ in eq.(\ref{Posterior}). 
They are determined so as to satisfy 
the following simultaneous 
fixed-point equations:
\begin{eqnarray}
{\mu}_{j{\to}i}({\xi})
& = &
{\frac{
{\displaystyle{{\sum_{{\zeta}{\in}{\cal{Q}}}}}}
{\exp}{\big (}{\frac{1}{2}}{\alpha}(u){\delta}_{{\xi},{\zeta}}{\big )}
g({\bm{d}}_{j}|{\zeta},{\bm{\Theta}})
{\displaystyle{
{\prod_{k{\in}{\partial}j{\backslash}\{i\}}}
}}
{\mu}_{k{\to}j}({\zeta})
}
{
{\displaystyle{{\sum_{{\zeta}{\in}{\cal{Q}}}}}}
{\displaystyle{{\sum_{{\zeta}'{\in}{\cal{Q}}}}}}
{\exp}{\big (}{\frac{1}{2}}{\alpha}(u){\delta}_{{\zeta}',{\zeta}}{\big )}
g({\bm{d}}_{j}|{\zeta},{\bm{\Theta}})
{\displaystyle{
{\prod_{k{\in}{\partial}j{\backslash}\{i\}}}
}}
{\mu}_{k{\to}j}({\zeta})
}
}
\nonumber\\
& &{\hspace{3.0cm}}
(i{\in}{\cal{V}},j{\in}{\partial}i,{\xi}{\in}{\cal{Q}}).
\label{LBPMessagePassingRuleB}
\end{eqnarray}
Here ${\partial}i$ denotes the pixels that are neighbours
of pixel $i$.
The left-hand sides in eqs.(\ref{Determination-u-posterior}),
(\ref{Determination-m-posterior})
and (\ref{Determination-C-posterior})
can be computed by means 
of eqs.(\ref{Determination-u-posteriorB}), 
(\ref{Determination-m-posteriorB}), and
 (\ref{LBPMessagePassingRuleB}).

The practical segmentation algorithm for
an observed image ${\bm{d}}$ 
is summarized as follows:
\begin{description}
\vspace{0.2500cm}
\item[{\hspace{0.250cm}}
{\framebox{\bf{Inference Algorithm 
for ${\widehat{u}}({\bm{d}})$, 
${\alpha}{\big (}{\widehat{u}}({\bm{d}}){\big )}$ 
and ${\widehat{\bm{\Theta}}}({\bm{d}})$}}}] 
\item[{\hspace{0.50cm}}Step 1]~Input the data ${\bm{d}}$. 
Set initial values for $u$, ${\cal{K}}$, 
${\bm{\Theta}}{\equiv}\{{\bm{m}}({\zeta}),
{\bm{C}}({\zeta})|{\zeta}{\in}{\cal{Q}}\}$ 
and 
$\{{\widehat{\mu}}_{j{\to}i}({\xi})|
i{\in}{\cal{V}},j{\in}{\partial}i,{\xi}{\in}{\cal{Q}}\}$, 
and $t \leftarrow 0$. 
\item[{\hspace{0.50cm}}Step 2]~
Set initial values for 
$\{{\widehat{\lambda}}_{j{\to}i}({\xi})|
i{\in}{\cal{V}},j{\in}{\partial}i,{\xi}{\in}{\cal{Q}}\}$ 
and repeat 
the following update rules 
until ${\cal{K}}$ and 
$\{{\widehat{\lambda}}_{j{\to}i}({\xi})|
i{\in}{\cal{V}},j{\in}{\partial}i,{\xi}{\in}{\cal{Q}}\}$ 
converge:
\begin{eqnarray}
{\lambda}_{j{\to}i}({\xi})
& \leftarrow &
{\frac{
{\displaystyle{{\sum_{{\zeta}{\in}{\cal{Q}}}}}}
{\exp}{\big (}{\frac{1}{2}}{\cal{K}}{\delta}_{{\xi},{\zeta}}{\big )}
{\displaystyle{
{\prod_{k{\in}{\partial}j{\backslash}\{i\}}}
}}
{\widehat{\lambda}}_{k{\to}j}({\zeta})
}
{
{\displaystyle{{\sum_{{\zeta}{\in}{\cal{Q}}}}}}
{\displaystyle{{\sum_{{\zeta}'{\in}{\cal{Q}}}}}}
{\exp}{\big (}{\frac{1}{2}}{\cal{K}}{\delta}_{{\zeta}',{\zeta}}{\big )}
{\displaystyle{
{\prod_{k{\in}{\partial}j{\backslash}\{i\}}}
}}
{\widehat{\lambda}}_{k{\to}j}({\zeta})
}
}
\nonumber\\
& &{\hspace{3.0cm}}
({\xi}{\in}{\cal{Q}},
~i{\in}{\cal{V}},~j{\in}{\partial}i),
\end{eqnarray}
\begin{eqnarray}
& &{\hspace{-2.0cm}}
{\cal{A}}_{i}
\leftarrow 
{\displaystyle{{\sum_{{\zeta}{\in}{\cal{Q}}}}}}
{\prod_{k{\in}{\partial}i}}
{\lambda}_{k{\to}i}({\zeta})
{\hspace{2.0mm}}
(i{\in}{\cal{V}}),
\end{eqnarray}
\begin{eqnarray}
& &{\hspace{-1.0cm}}
{\cal{A}}_{\{i,j\}}
\leftarrow 
{\displaystyle{{\sum_{{\zeta}{\in}{\cal{Q}}}}}}
{\displaystyle{{\sum_{{\zeta}'{\in}{\cal{Q}}}}}}
{\Big (}
{\prod_{k{\in}{\partial}i{\backslash}\{j\}}}
{\lambda}_{k{\to}i}({\zeta})
{\Big )}
{\exp}{\Big (}{\frac{1}{2}}{\cal{K}}{\delta}_{{\zeta},{\zeta}'}{\Big )}
\nonumber\\
& &{\hspace{1.0cm}}{\times}
{\Big (}
{\prod_{k{\in}{\partial}j{\backslash}\{i\}}}
{\lambda}_{k{\to}j}({\zeta}')
{\Big )}~
(\{i,j\}{\in}{\cal{E}}),
\end{eqnarray}
\begin{eqnarray}
{\cal{K}}
& \leftarrow & 
{\cal{K}}{\times}
{\Bigg (}
{\frac{1}{u|{\cal{E}}|}}
{\sum_{\{i,j\}{\in}{\cal{E}}}}
{\frac{1}{{\cal{A}}_{\{i,j\}}}}
{\displaystyle{{\sum_{{\zeta}{\in}{\cal{Q}}}}}}
{\displaystyle{{\sum_{{\zeta}'{\in}{\cal{Q}}}}}}
{\big (}1-{\delta}_{{\zeta},{\zeta}'}{\big )}
\nonumber\\
& &{\hspace{1.0cm}}
{\times}
{\Big (}
{\prod_{k{\in}{\partial}i{\backslash}\{j\}}}
{\lambda}_{k{\to}i}({\zeta})
{\Big )}
{\exp}{\Big (}{\frac{1}{2}}{\cal{K}}{\delta}_{{\zeta},{\zeta}'}{\Big )}
{\Big (}
{\prod_{k{\in}{\partial}j{\backslash}\{i\}}}
{\lambda}_{k{\to}j}({\zeta}')
{\Big )}
{\Bigg )}^{1/4}.
\end{eqnarray}
\begin{eqnarray}
{\widehat{\lambda}}_{j{\to}i}({\xi})
& \leftarrow &
{\lambda}_{j{\to}i}({\xi})~
({\xi}{\in}{\cal{Q}},
~i{\in}{\cal{V}},~j{\in}{\partial}i).
\end{eqnarray}
\item[{\hspace{0.50cm}}Step 3]~Update 
${\bm{\Theta}}$,
$\{{\mu}_{j{\to}i}({\xi})|{\xi}{\in}{\cal{Q}},i{\in}{\cal{V}},j{\in}{\partial}i\}$ 
and $u$ according to the following rules:
\begin{eqnarray}
{\mu}_{j{\to}i}({\xi})
& \leftarrow & 
{\frac{
{\displaystyle{{\sum_{{\zeta}{\in}{\cal{Q}}}}}}
{\exp}{\big (}{\frac{1}{2}}{\alpha}(u){\delta}_{{\xi},{\zeta}}{\big )}
g({\bm{d}}_{j}|{\zeta},{\bm{\Theta}})
{\displaystyle{
{\prod_{k{\in}{\partial}j{\backslash}\{i\}}}
}}
{\widehat{\mu}}_{k{\to}j}({\zeta})
}
{
{\displaystyle{{\sum_{{\zeta}{\in}{\cal{Q}}}}}}
{\displaystyle{{\sum_{{\zeta}'{\in}{\cal{Q}}}}}}
{\exp}{\big (}{\frac{1}{2}}{\alpha}(u){\delta}_{{\zeta}',{\zeta}}{\big )}
g({\bm{d}}_{j}|{\zeta},{\bm{\Theta}})
{\displaystyle{
{\prod_{k{\in}{\partial}j{\backslash}\{i\}}}
}}
{\widehat{\mu}}_{k{\to}j}({\zeta})
}
}
\nonumber\\
& &{\hspace{3.0cm}}
({\xi}{\in}{\cal{Q}},
i{\in}{\cal{V}},j{\in}{\partial}i),
\end{eqnarray}
\begin{eqnarray}
{\cal{B}}_{i}
& \leftarrow & 
{\displaystyle{{\sum_{{\zeta}{\in}{\cal{Q}}}}}}
g({\bm{d}}_{i}|{\zeta},{\bm{\Theta}})
{\prod_{k{\in}{\partial}i}}
{\mu}_{k{\to}i}({\zeta})
{\hspace{2.0mm}}
(i{\in}{\cal{V}}),
\end{eqnarray}
\begin{eqnarray}
{\cal{B}}_{\{i,j\}}
& \leftarrow &
{\displaystyle{{\sum_{{\zeta}{\in}{\cal{Q}}}}}}
{\displaystyle{{\sum_{{\zeta}'{\in}{\cal{Q}}}}}}
{\Big (}
{\prod_{k{\in}{\partial}i{\backslash}\{j\}}}
{\mu}_{k{\to}i}({\zeta})
{\Big )}
g({\bm{d}}_{i}|{\zeta},{\bm{\Theta}})
{\exp}{\big (}{\frac{1}{2}}{\alpha}(u){\delta}_{{\zeta},{\zeta}'}{\big )}
\nonumber\\
& &{\hspace{1.0cm}}{\times}
g({\bm{d}}_{j}|{\zeta}',{\bm{\Theta}})
{\Big (}
{\prod_{k{\in}{\partial}j{\backslash}\{i\}}}
{\mu}_{k{\to}j}({\zeta}')
{\Big )}
~(\{i,j\}{\in}{\cal{E}}),
\end{eqnarray}
\begin{eqnarray}
{\bm{m}}({\xi})
& \leftarrow & 
{\frac{
{\displaystyle{{\sum_{i{\in}{\cal{V}}}}}}
{\displaystyle{{\frac{1}{{\cal{B}}_{i}}}}}
{\bm{d}}_{i}
g({\bm{d}}_{i}|{\xi},{\bm{\Theta}})
{\Big (}
{\displaystyle{{\prod_{k{\in}{\partial}i}}}}
{\mu}_{k{\to}i}({\xi})
{\Big )}
}{
{\displaystyle{{\sum_{i{\in}{\cal{V}}}}}}
{\displaystyle{{\frac{1}{{\cal{B}}_{i}}}}}
g({\bm{d}}_{i}|{\xi},{\bm{\Theta}})
{\Big (}
{\displaystyle{{\prod_{k{\in}{\partial}i}}}}
{\mu}_{k{\to}i}({\xi})
{\Big )}
}},
\end{eqnarray}
\begin{eqnarray}
{\bm{C}}({\xi})
& \leftarrow & 
{\frac{
{\displaystyle{{\sum_{i{\in}{\cal{V}}}}}}
{\displaystyle{{\frac{1}{{\cal{B}}_{i}}}}}
{\big (}{\bm{d}}_{i}-{\bm{m}}({\xi}){\big )}
{\big (}{\bm{d}}_{i}-{\bm{m}}({\xi}){\big )}^{\rm{T}}
g({\bm{d}}_{i}|{\xi},{\bm{\Theta}})
{\Big (}
{\displaystyle{{\prod_{k{\in}{\partial}i}}}}
{\mu}_{k{\to}i}({\xi})
{\Big )}
}{
{\displaystyle{{\sum_{i{\in}{\cal{V}}}}}}
{\displaystyle{{\frac{1}{{\cal{B}}_{i}}}}}
g({\bm{d}}_{i}|{\xi},{\bm{\Theta}})
{\Big (}
{\displaystyle{{\prod_{k{\in}{\partial}i}}}}
{\mu}_{k{\to}i}({\xi})
{\Big )}
}},
\end{eqnarray}
\begin{eqnarray}
{\bm{\Theta}}
& \leftarrow & 
\{{\bm{m}}({\xi}),{\bm{C}}({\xi})|{\xi}{\in}{\cal{Q}}\},
\end{eqnarray}
\begin{eqnarray}
u & \leftarrow & 
{\frac{1}{|{\cal{E}}|}}
{\sum_{\{i,j\}{\in}{\cal{E}}}}
{\Bigg (}
{\frac{1}{{\cal{B}}_{\{i,j\}}}}
{\displaystyle{{\sum_{{\zeta}{\in}{\cal{Q}}}}}}
{\displaystyle{{\sum_{{\zeta}'{\in}{\cal{Q}}}}}}
{\big (}1-{\delta}_{{\zeta},{\zeta}'}{\big )}
{\Big (}
{\prod_{k{\in}{\partial}i{\backslash}\{j\}}}
{\mu}_{k{\to}i}({\zeta})
{\Big )}
\nonumber\\
& &{\hspace{1.0cm}}{\times}
g({\bm{d}}_{i}|{\zeta},{\bm{\Theta}})
{\exp}{\big (}{\frac{1}{2}}{\alpha}(u){\delta}_{{\zeta},{\zeta}'}{\big )}
g({\bm{d}}_{j}|{\zeta}',{\bm{\Theta}})
{\Big (}
{\prod_{k{\in}{\partial}j{\backslash}\{i\}}}
{\mu}_{k{\to}j}({\zeta}')
{\Big )}
{\Bigg )}.
\end{eqnarray}
\begin{eqnarray}
{\widehat{\mu}}_{j{\to}i}({\xi})
& \leftarrow &
{\mu}_{j{\to}i}({\xi})~
({\xi}{\in}{\cal{Q}},
~i{\in}{\cal{V}},~j{\in}{\partial}i).
\end{eqnarray}
\item[{\hspace{0.50cm}}Step 4]~
Output the following quantities:
\begin{eqnarray}
& &t \leftarrow t+1,
{\hspace{2.0mm}}
{\widehat{u}}({\bm{d}}){\leftarrow}u,
{\hspace{2.0mm}}
{\alpha}({\widehat{u}}({\bm{d}})){\leftarrow}{\cal{K}},
\\
& &{\widehat{\bm{\Theta}}}({\bm{d}})
\equiv 
\{{\widehat{\bm{m}}}({\xi},{\bm{d}}),{\widehat{\bm{C}}}({\xi},{\bm{d}})
|{\xi}{\in}{\cal{Q}}\}
\leftarrow {\bm{\Theta}},
\\
& &{\widehat{a}}_{i}({\bm{d}})
\leftarrow 
{\arg}{\max_{{\zeta}{\in}{\cal{Q}}}}
g({\bm{d}}_{i}|{\zeta},{\bm{\Theta}})
{\prod_{k{\in}{\partial}i}}
{\mu}_{k{\to}i}({\zeta})
{\hspace{2.0mm}}
(i{\in}{\cal{V}}),
\end{eqnarray}
and stop if ${\widehat{u}}({\bm{d}})$ and 
${\widehat{\bm{\Theta}}}({\bm{d}})$ converge.
Go to {\bf{Step 2}} otherwise.
\end{description}

We use six test images, as shown in Figs.{\ref{Figure03}}(a)-(f), 
where three images are from the Berkeley Segmentation 
Data Set 500 (BSDS500){\cite{BSDS500,ArbelaezMaireFowlkesMalik2011}} 
and the other three images are 
from the image database of Signal and Image Processing Institute, 
University of Southern California (SPIP-USC){\cite{SPIP-USC}} 
to demonstrate the effectiveness of our method. 
The processes of the proposed hyperparameter estimation for the images
in Fig.{\ref{Figure03}}(a)-(f) are plotted 
in Figs.{\ref{Figure04}}(a)-(f) and {\ref{Figure05}}(a)-(f) 
under $q=5$ and $q=8$, respectively. 
The solid circles in Figs.{\ref{Figure04}} and {\ref{Figure05}} 
correspond to
$({\alpha}({\widehat{u}}({\bm{d}})),{\widehat{u}}({\bm{d}}))$ in Step 4, and 
the solid lines are ${\alpha}(u)$ for various values of $u$ 
and are also given in Fig.{\ref{Figure02}}.
In Table {\ref{Table01}}, we show the estimates 
${\widehat{u}}({\bm{d}})$ and ${\alpha}({\widehat{u}}({\bm{d}}))$ 
in the cases of $q=5$ and $q=8$ 
for the images ${\bm{d}}$ in Fig.{\ref{Figure03}}. 
The segmentation results 
for the test images ${\bm{d}}$  
in Fig.{\ref{Figure03}} are shown 
in Figs.{\ref{Figure06}} and {\ref{Figure07}} for $q=5$ and $q=8$,
where the results are represented
as color images 
${\Big (}
{\widehat{\bm{m}}}{\big (}{\widehat{a}}_{1}({\bm{d}}),{\bm{d}}{\big )},
{\cdots},
{\widehat{\bm{m}}}{\big (}{\widehat{a}}_{|{\cal{V}}|}({\bm{d}}),{\bm{d}}{\big )}
{\Big )}^{\rm{T}}$ in terms of 
the average vectors ${\bm{m}}({\xi},{\bm{d}})$ (${\xi}{\in}{\cal{Q}}$) 
and the estimate of labeling ${\widehat{a}}({\bm{d}})$.
\begin{figure}
\begin{center}
{\bf{(a)}}
{\hspace{4.250cm}}
{\bf{(b)}}
\\
\includegraphics[height=3.0cm, bb = 0 0 288 192]{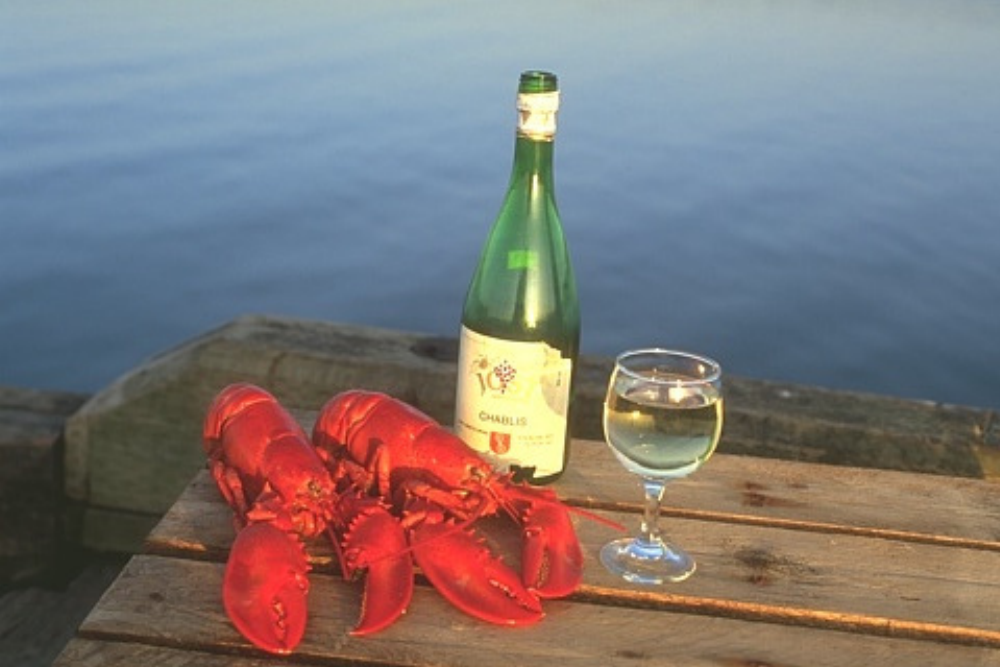}
{\hspace{5.0mm}}
\includegraphics[height=3.0cm, bb = 0 0 1154 770]{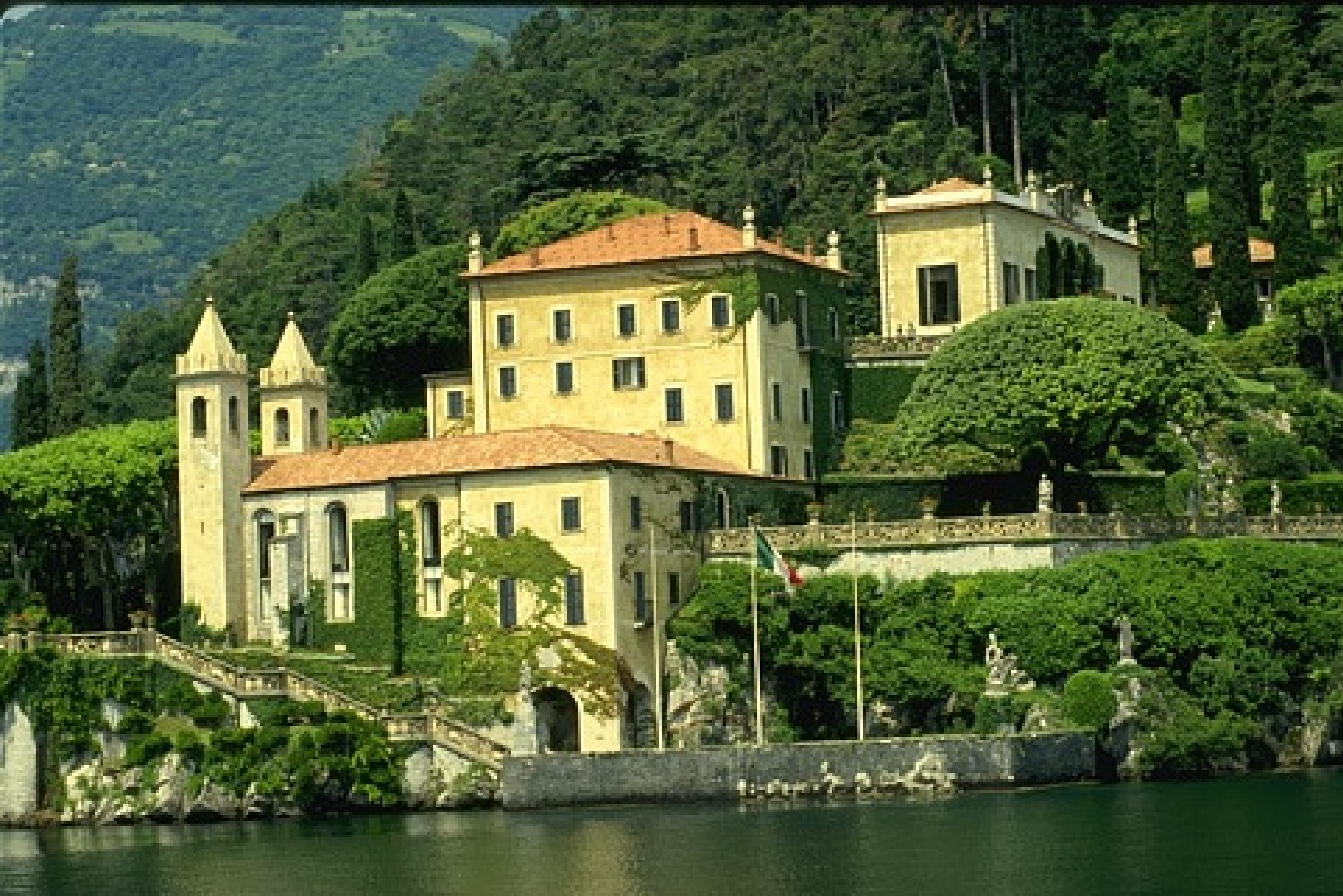}
\\
{\bf{(c)}}
{\hspace{4.250cm}}
{\bf{(d)}}
\\
\includegraphics[height=3.0cm, bb = 0 0 288.6 192.6]{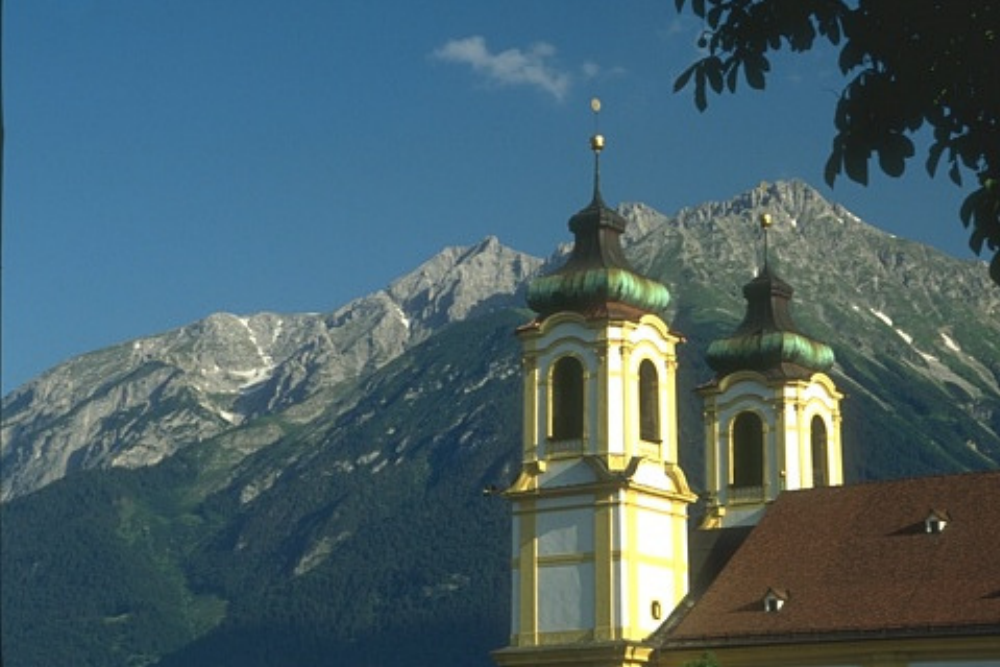}
{\hspace{5.0mm}}
\includegraphics[height=3.0cm, bb = 0 0 288.6 192.6]{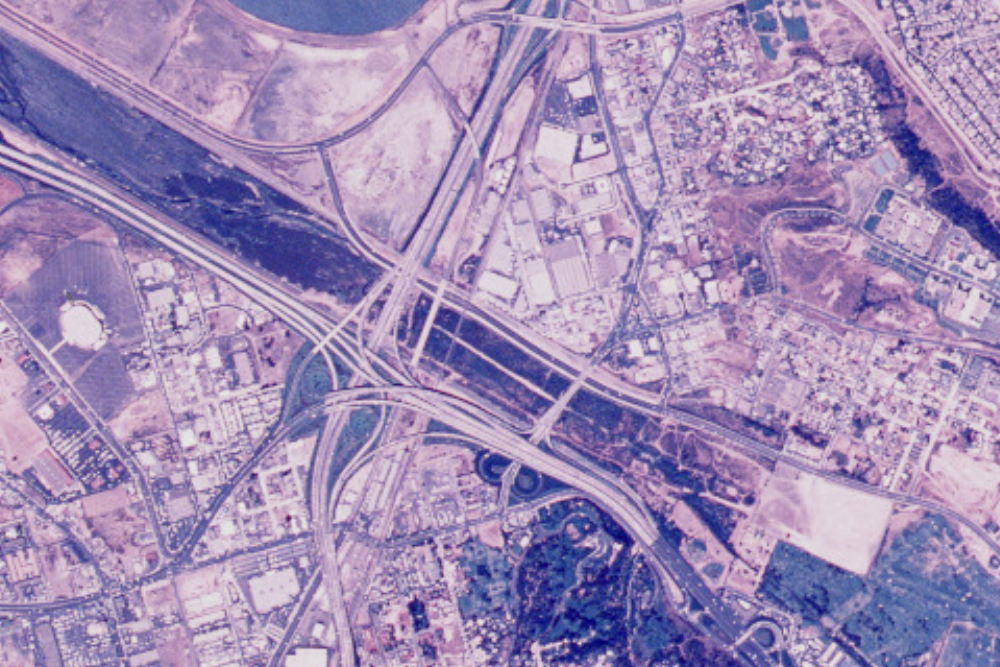}
\\
{\bf{(e)}}
{\hspace{4.250cm}}
{\bf{(f)}}
\\
\includegraphics[height=3.0cm, bb = 0 0 288.6 192.6]{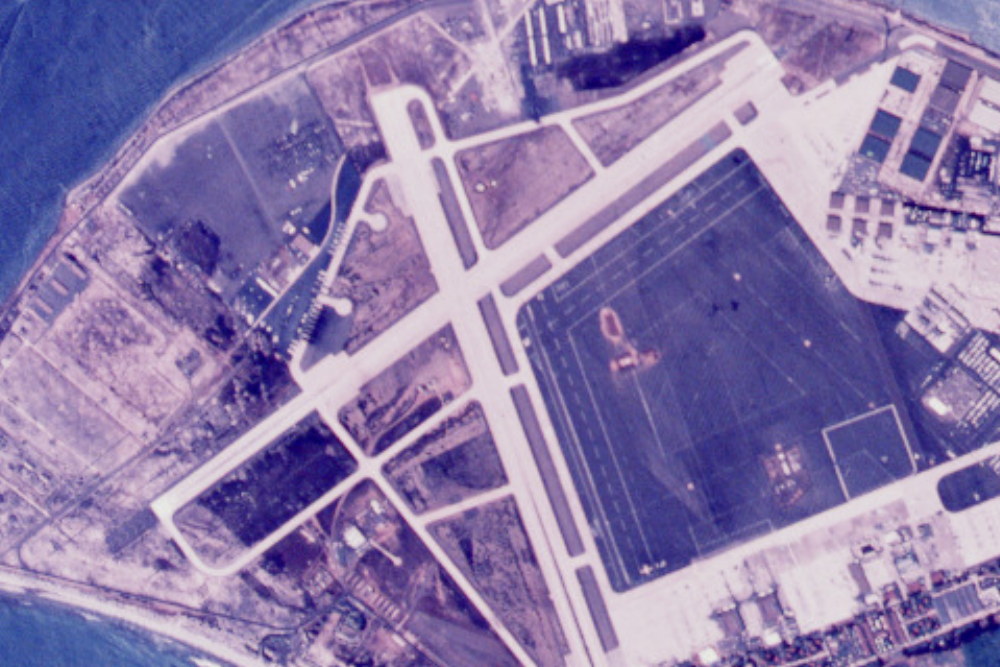}
{\hspace{5.0mm}}
\includegraphics[height=3.0cm, bb = 0 0 288 192]{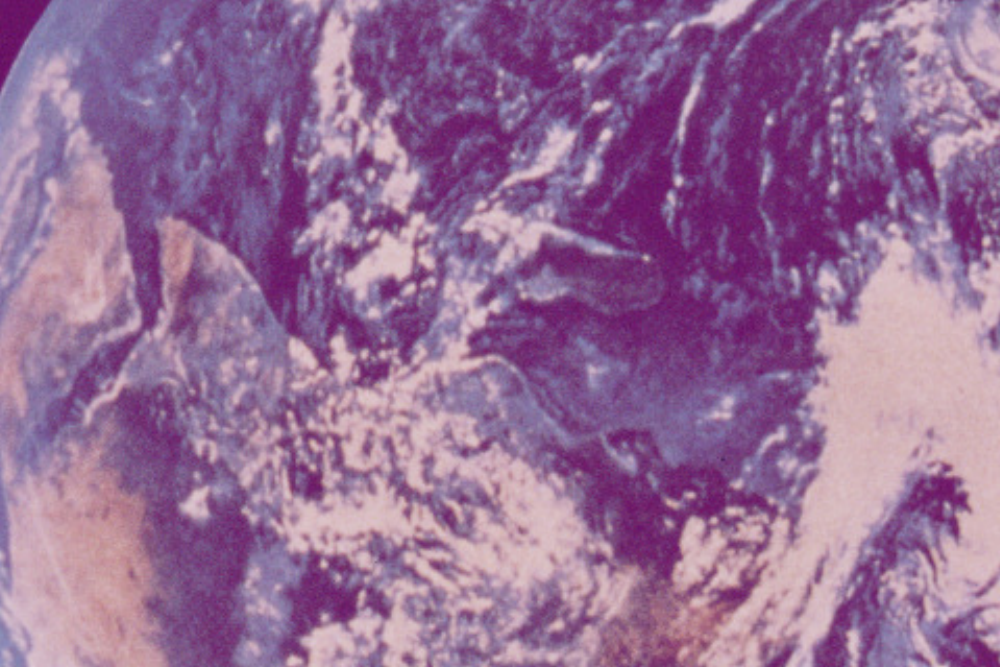}
\end{center}
\caption{(a)-(c) Three test images 
in the Berkeley Segmentation Data Set 500 
(BSDS500){\cite{BSDS500,ArbelaezMaireFowlkesMalik2011}}. 
(d)-(f) Three test images 
from the database of Signal and Image Processing Institute, 
University of Southern California (SPIP-USC){\cite{SPIP-USC}}. 
Each color image is represented by
${\bm{d}}=({\bm{d}}_{1},{\bm{d}}_{2},{\cdots},
{\bm{d}}_{|{\cal{V}}|})^{\rm{T}}$,
where 
${\bm{d}}_{i}=(d_{i}^{\rm{R}},d_{i}^{\rm{G}},d_{i}^{\rm{B}})^{\rm{T}}$}
\label{Figure03}
\end{figure}
\begin{figure}
\begin{center}
\includegraphics[height=8.0cm, bb = 0 0 798 554]{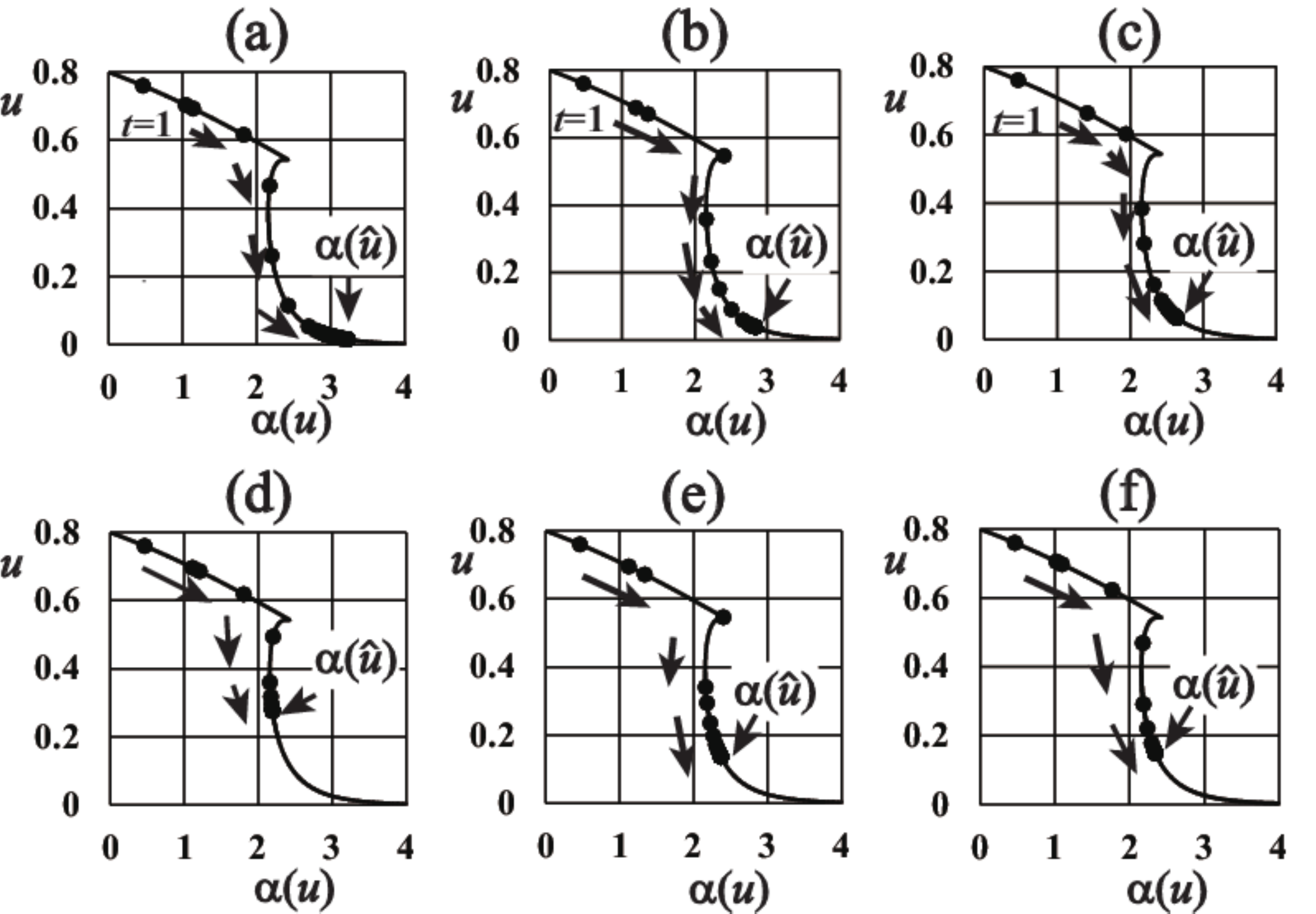}
\end{center}
\caption{Hyperparameter estimation process 
by using our proposed inference algorithm of {\S}3 for $q=5$. 
The solid circles in (a)-(f) are 
$({\alpha}({\widehat{u}}({\bm{d}})),{\widehat{u}}({\bm{d}}))$ 
at $t=1,2,3,{\cdots}$ in Step 4 
for the images ${\bm{d}}$ 
in Fig.{\ref{Figure03}}(a)-(f), respectively. 
Our estimation process in the proposed inference algorithm almost converges 
within $t \ge 30$ for each ${\bm{d}}$. 
The solid lines are ${\alpha}(u)$ for various values of $u$ 
and are also given in Fig.{\ref{Figure02}}(a).
}
\label{Figure04}
\end{figure}
\begin{figure}
\begin{center}
\includegraphics[height=8.0cm, bb = 0 0 793 563]{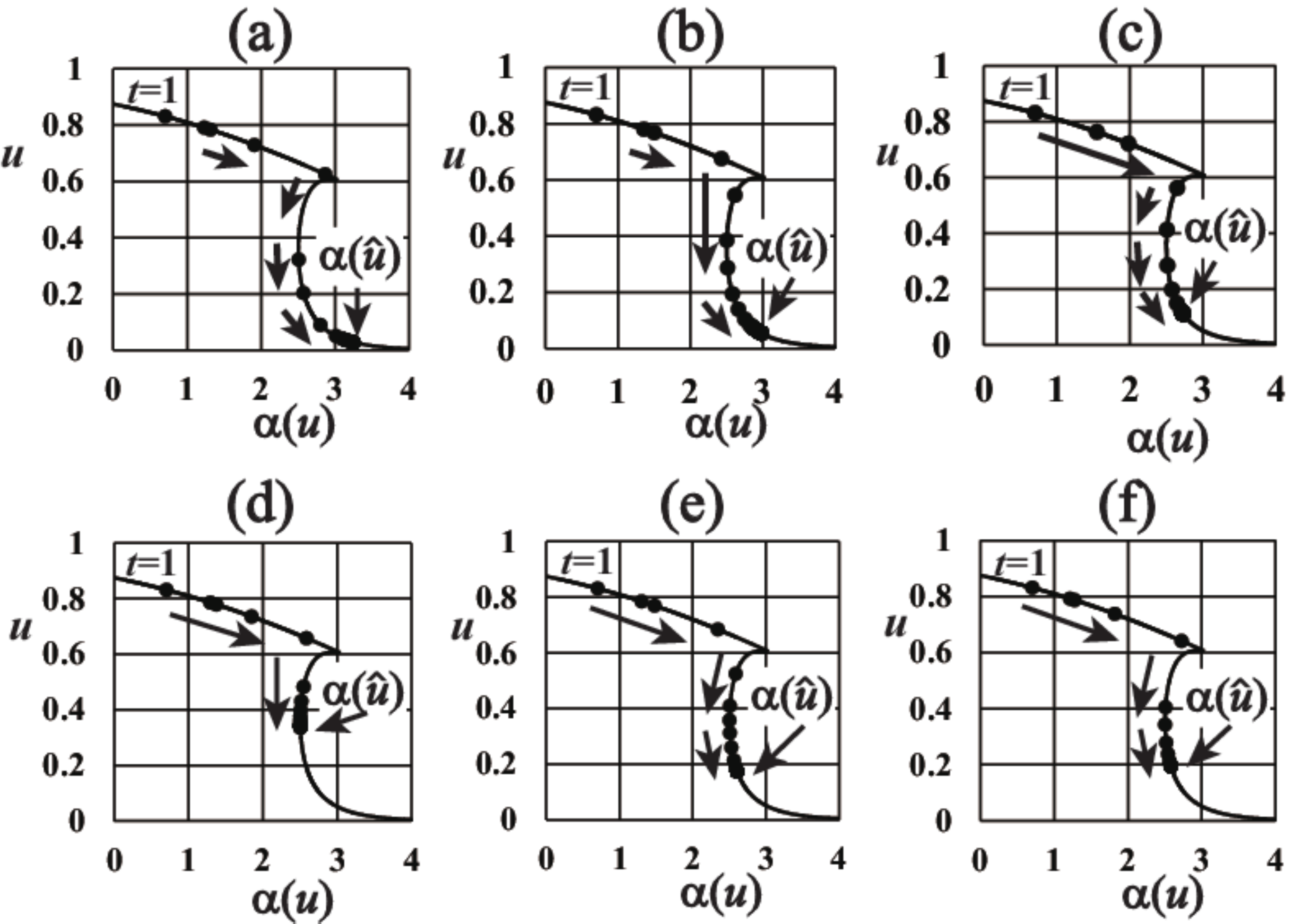}
\end{center}
\caption{Hyperparameter estimation process 
by using our proposed inference algorithm in {\S}3 for $q=8$. 
The solid circles in (a)-(f) are 
$({\alpha}({\widehat{u}}({\bm{d}})),{\widehat{u}}({\bm{d}}))$ 
at $t=1,2,3,{\cdots}$ in Step 4 
for the images ${\bm{d}}$ 
in Fig.{\ref{Figure03}}(a)-(f), respectively. 
Our estimation process in the proposed inference algorithm almost converges 
within $t \ge 30$ for each ${\bm{d}}$. 
The solid lines are ${\alpha}(u)$ for various values of $u$ 
and are also given in Fig.{\ref{Figure02}}(b).
}
\label{Figure05}
\end{figure}
\begin{figure}
\begin{center}
{\bf{(a)}}
{\hspace{4.250cm}}
{\bf{(b)}}
\\
\includegraphics[height=3.0cm, bb = 0 0 288 192]{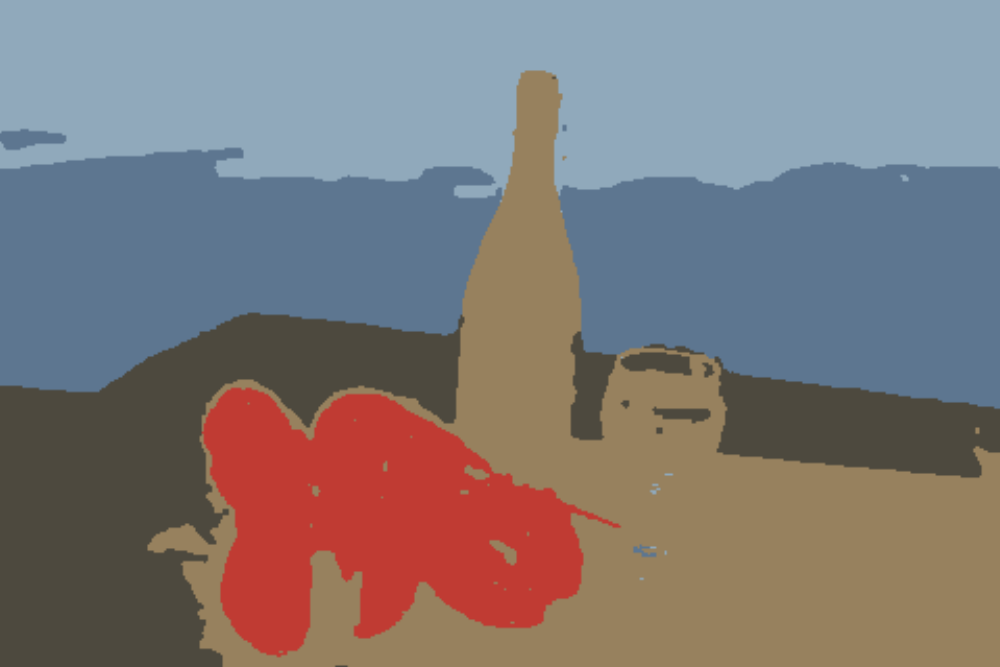}
{\hspace{5.0mm}}
\includegraphics[height=3.0cm, bb = 0 0 288 192]{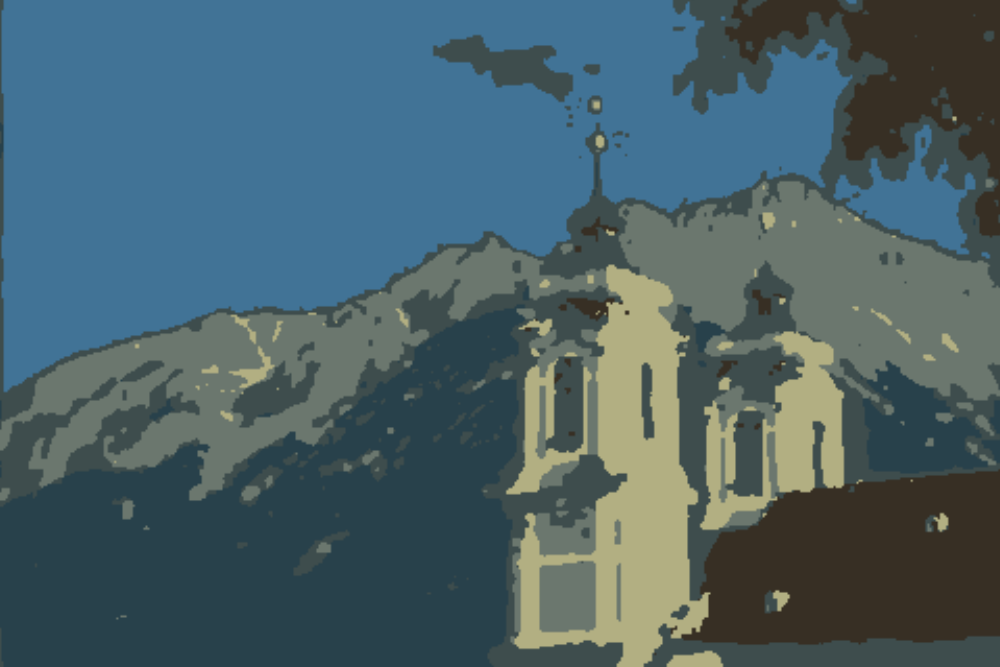}
\\
{\bf{(c)}}
{\hspace{4.250cm}}
{\bf{(d)}}
\\
\includegraphics[height=3.0cm, bb = 0 0 288 192]{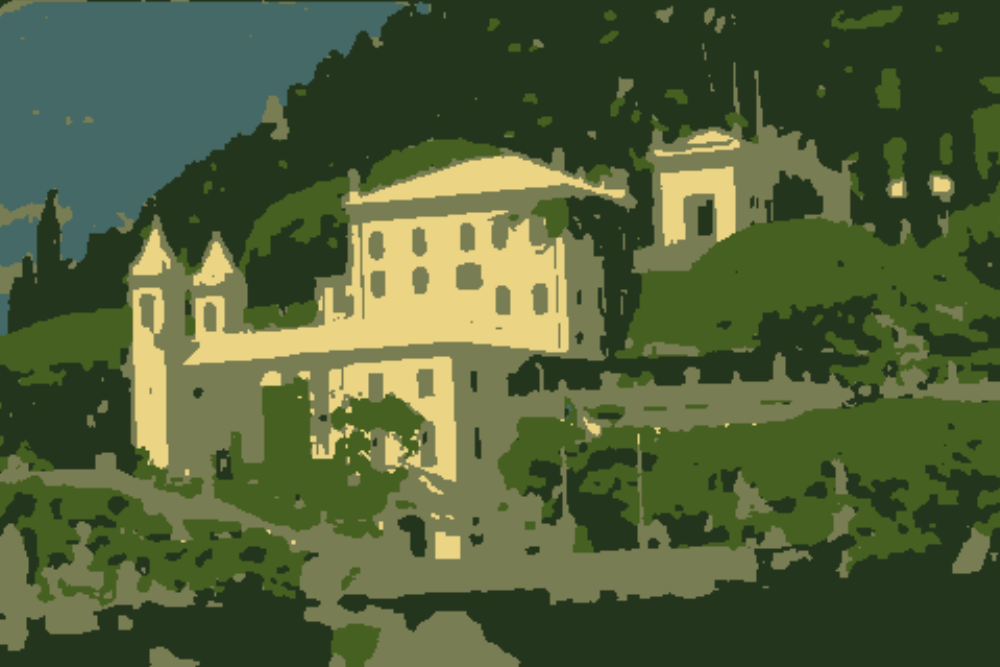}
{\hspace{5.0mm}}
\includegraphics[height=3.0cm, bb = 0 0 288 192]{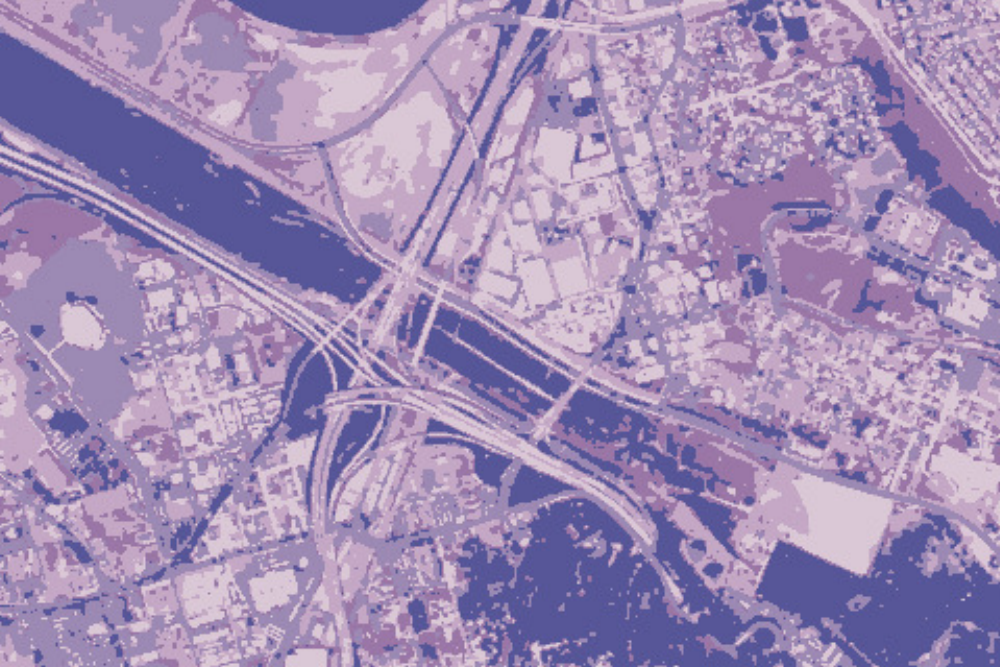}
\\
{\bf{(e)}}
{\hspace{4.250cm}}
{\bf{(f)}}
\\
\includegraphics[height=3.0cm, bb = 0 0 288 192]{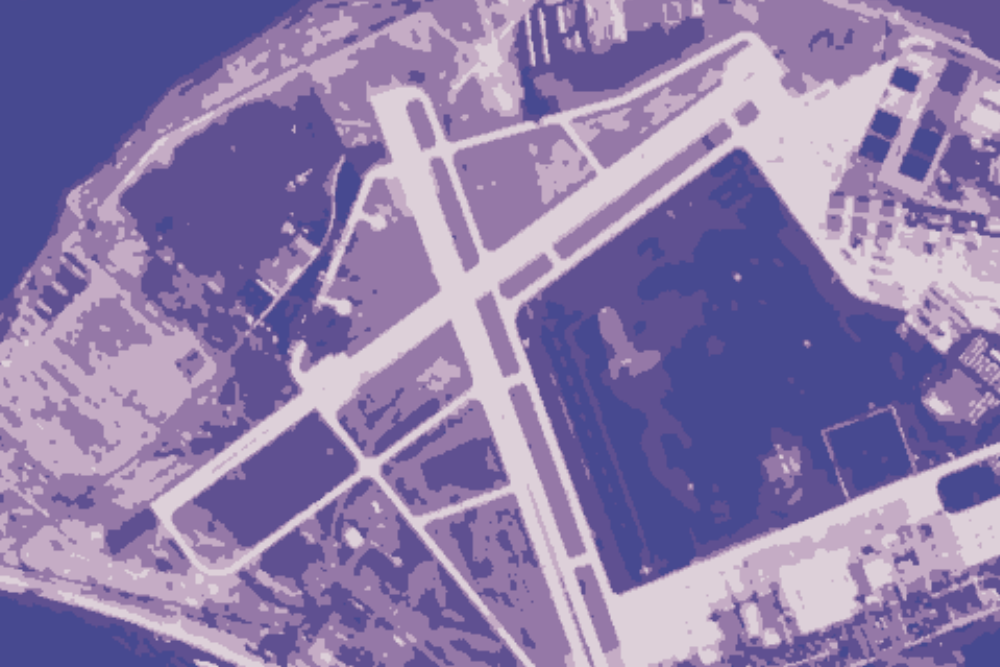}
{\hspace{5.0mm}}
\includegraphics[height=3.0cm, bb = 0 0 288 192]{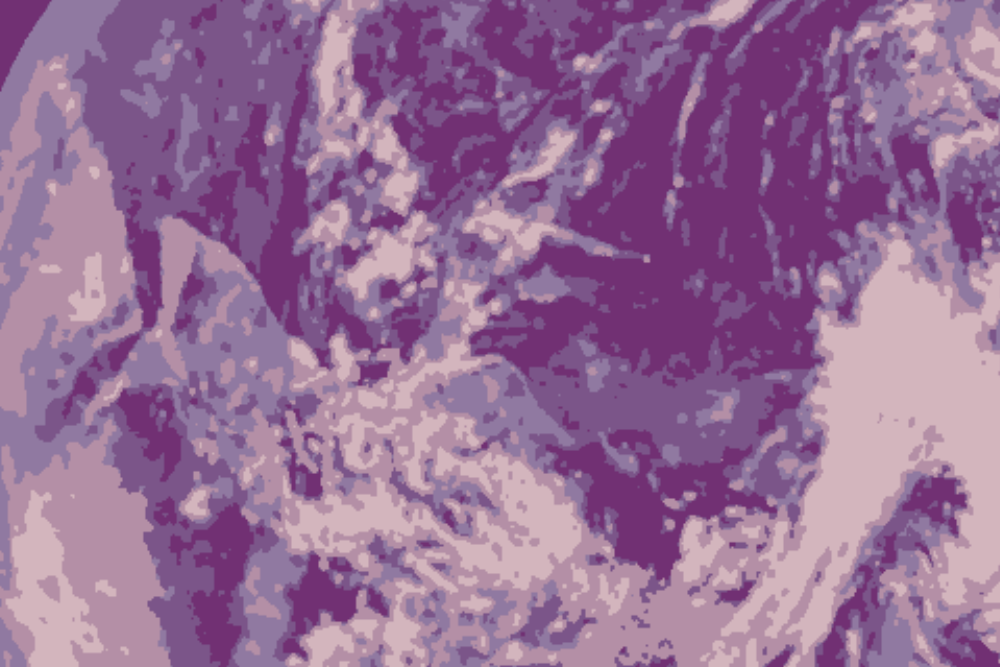}
\end{center}
\caption{Segmentation ${\bm{\widehat{a}}}({\bm{d}})
={\big (}{\widehat{a}}_{1}({\bm{d}}),{\widehat{a}}_{2}({\bm{d}}),
{\cdots},{\widehat{a}}_{|{\cal{V}}|}({\bm{d}}){\big )}^{\rm{T}}$ 
by using the proposed algorithm based 
on our conditional maximum entropy framework 
and the loopy belief propagation of {\S}3 
in the case of $q=5$.
The results in (a)-(f) are shown with the color 
${\widehat{\bm{m}}}{\big (}{\widehat{a}}_{i}({\bm{d}}),{\bm{d}}{\big )}$ 
at each pixel $i$ 
for the observed images ${\bm{d}}$ 
in Fig.{\ref{Figure03}}.
}
\label{Figure06}
\end{figure}
\begin{figure}
\begin{center}
{\bf{(a)}}
{\hspace{4.250cm}}
{\bf{(b)}}
\\
\includegraphics[height=3.0cm, bb = 0 0 288 192]{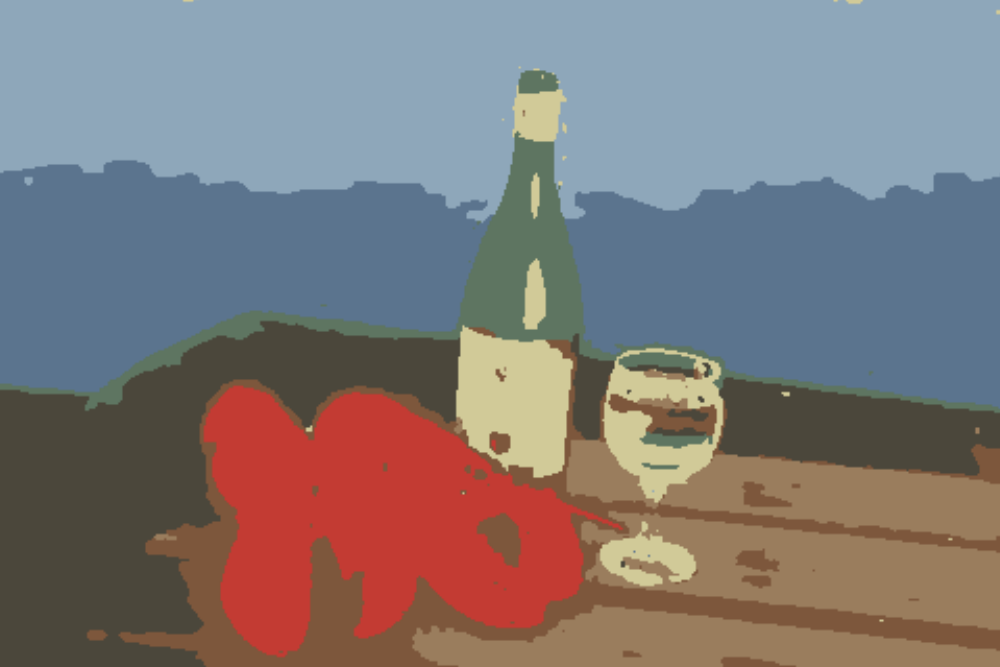}
{\hspace{5.0mm}}
\includegraphics[height=3.0cm, bb = 0 0 288 192]{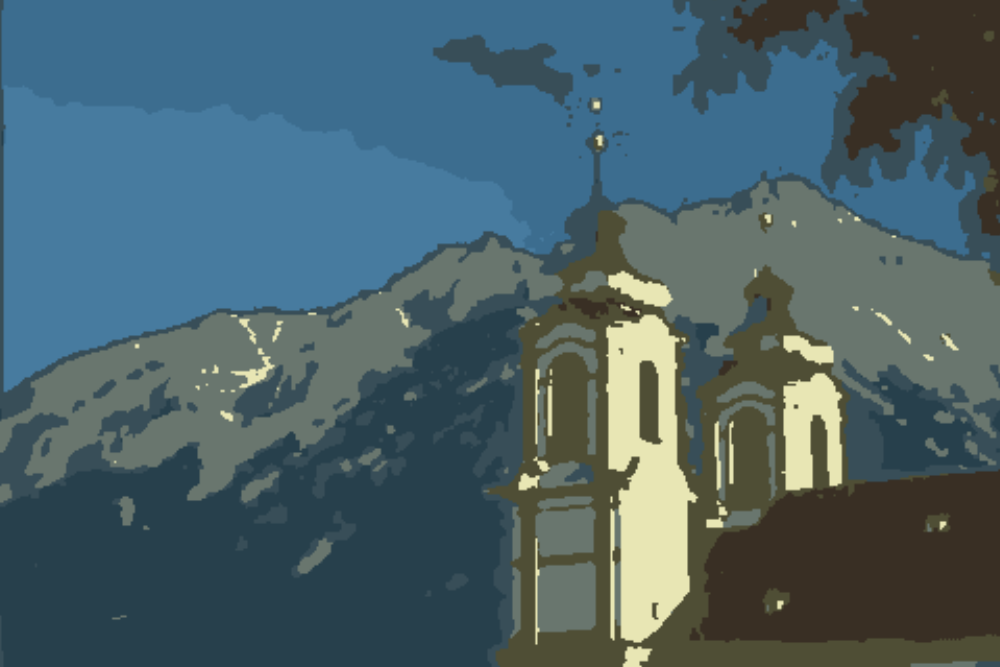}
\\
{\bf{(c)}}
{\hspace{4.250cm}}
{\bf{(d)}}
\\
\includegraphics[height=3.0cm, bb = 0 0 288 192]{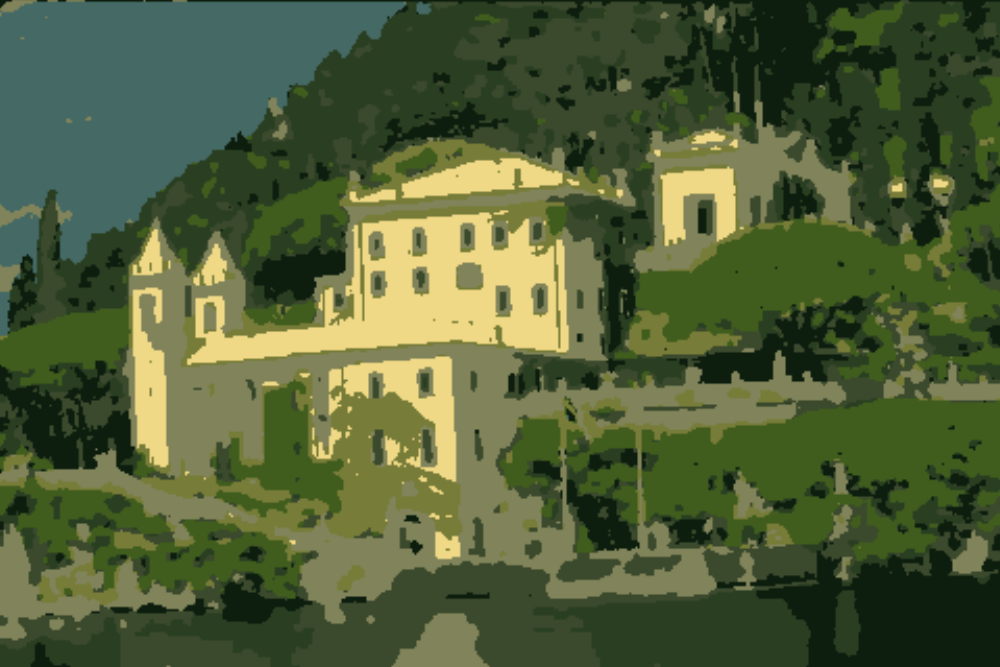}
{\hspace{5.0mm}}
\includegraphics[height=3.0cm, bb = 0 0 288 192]{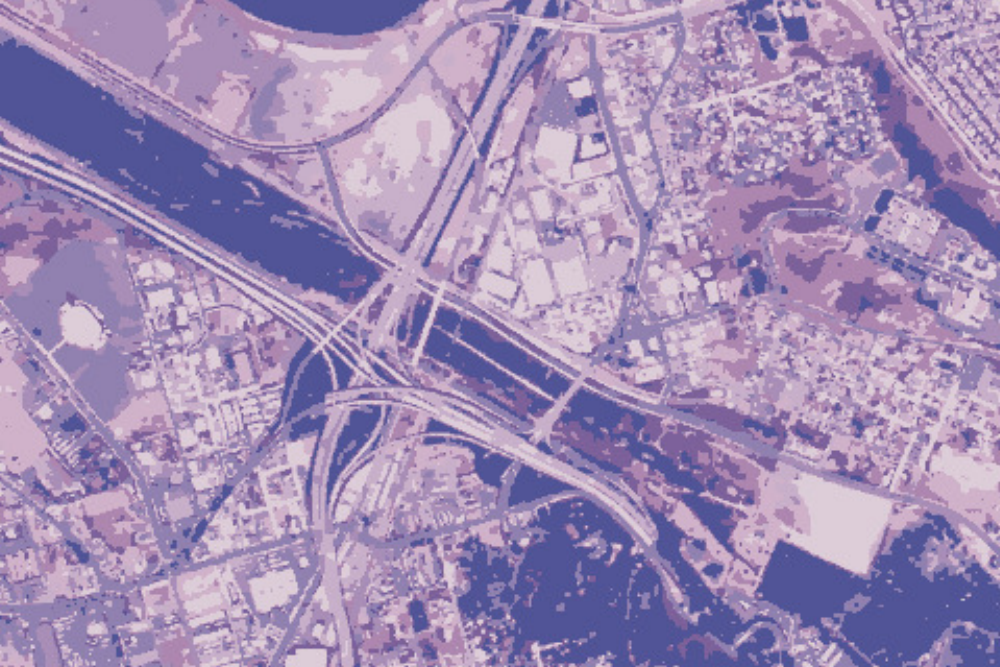}
\\
{\bf{(e)}}
{\hspace{4.250cm}}
{\bf{(f)}}
\\
\includegraphics[height=3.0cm, bb = 0 0 288 192]{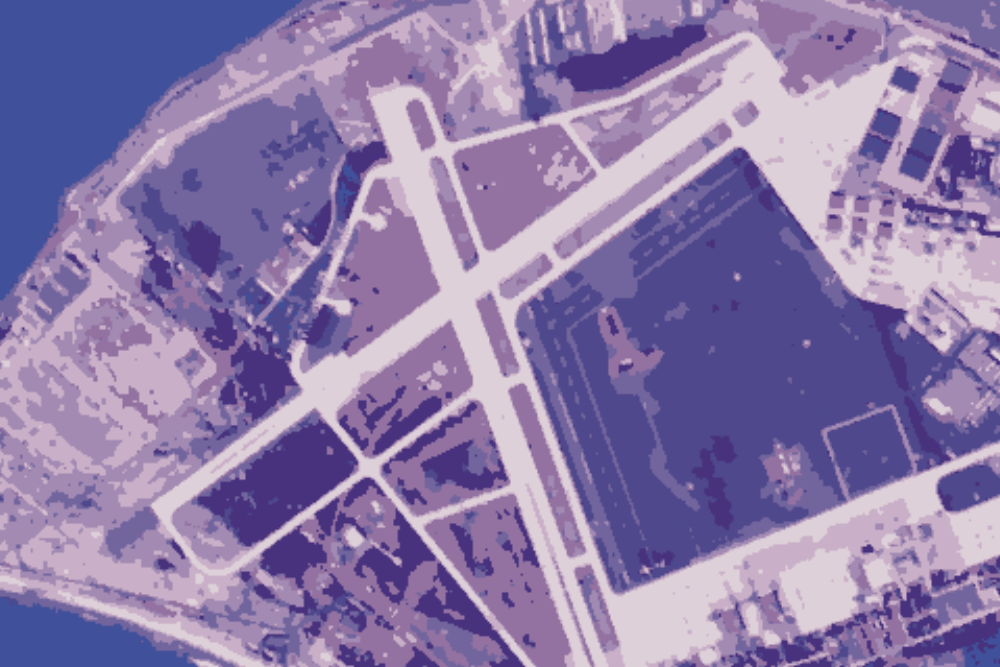}
{\hspace{5.0mm}}
\includegraphics[height=3.0cm, bb = 0 0 288 192]{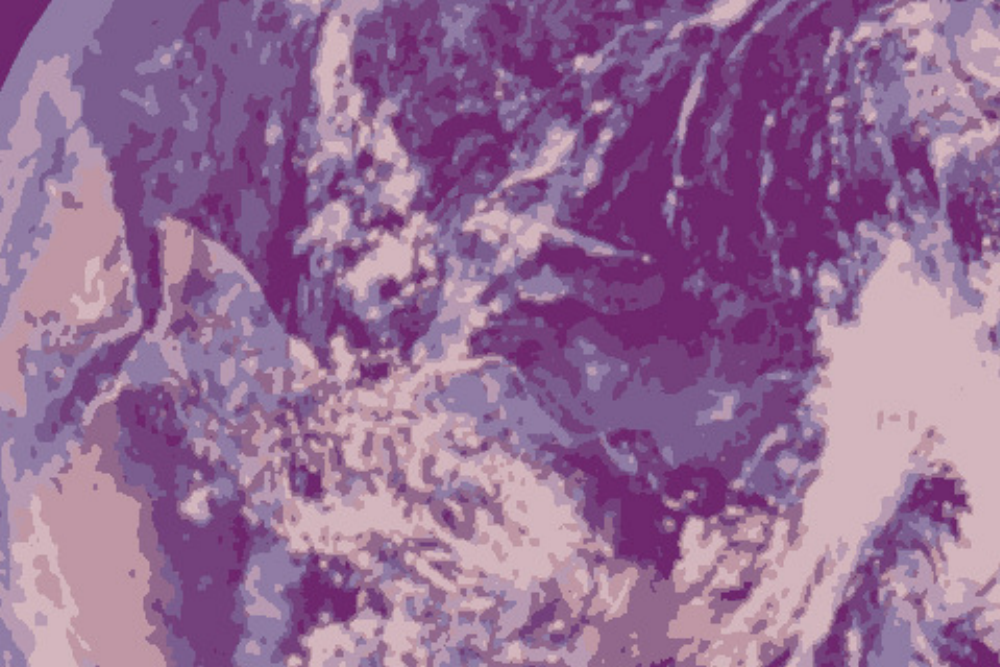}
\end{center}
\caption{Segmentation ${\bm{\widehat{a}}}({\bm{d}})
={\big (}{\widehat{a}}_{1}({\bm{d}}),{\widehat{a}}_{2}({\bm{d}}),
{\cdots},{\widehat{a}}_{|{\cal{V}}|}({\bm{d}}){\big )}^{\rm{T}}$ 
by using the proposed algorithm based 
on our conditional maximum entropy framework 
and the loopy belief propagation of {\S}3 
in the case of $q=8$.
The results in (a)-(f) are shown 
for the observed images ${\bm{d}}$ 
in Fig.{\ref{Figure03}}.
}
\label{Figure07}
\end{figure}
\begin{table}
\caption{Estimates of hyperparameters ${\widehat{u}}({\bm{d}})$ 
and ${\alpha}({\widehat{u}}({\bm{d}}))$ 
by using the proposed algorithm described in section 3 
for each observed image ${\bm{d}}$.
(a)~$q=5$. (b)~$q=8$.
Here $K_{\rm{C}}$'s 
are the first order transition points 
of the $q$-state Potts model 
in the loopy belief propagation and 
are $2.1972$ and $2.5871$ for $q=5$ and $q=8$, respectively.
}
\begin{center}
\hspace{5.0mm}
{\bf{(a)}} 
\begin{tabular}{|c|c|c|c|c|}
\hline
 ${\bm{d}}$ 
 & Fig.{\ref{Figure03}}(a)     & Fig.{\ref{Figure03}}(b)     & Fig.{\ref{Figure03}}(c)     \\
\hline
 ${\widehat{u}}({\bm{d}})$ 
 & $0.0155$                       & $0.0382$                       & $0.0631$                       \\
\hline
 ${\alpha}{\big (}{\widehat{u}}({\bm{d}}){\big )}$ 
 & $3.2218(>K_{\rm{C}})$          & $2.8367(>K_{\rm{C}})$          & $2.6397(>K_{\rm{C}})$          \\
\hline
\hline
 ${\bm{d}}$ 
 & Fig.{\ref{Figure03}}(d)     & Fig.{\ref{Figure03}}(e)     & Fig.{\ref{Figure03}}(f)     \\
\hline
 ${\widehat{u}}({\bm{d}})$ 
 & $0.2775$                       & $0.1440$                       & $0.1496$                       \\
\hline
 ${\alpha}{\big (}{\widehat{u}}({\bm{d}}){\big )}$ 
 & $2.1932(<K_{\rm{C}})$          & $2.3559(>K_{\rm{C}})$          & $2.3444(>K_{\rm{C}})$          \\
\hline
\end{tabular}
\end{center}
\begin{center}
\hspace{5.0mm}
{\bf{(b)}} 
\begin{tabular}{|c|c|c|c|c|}
\hline
 ${\bm{d}}$ 
 & Fig.{\ref{Figure03}}(a)     & Fig.{\ref{Figure03}}(b)     & Fig.{\ref{Figure03}}(c)     \\
\hline
 ${\widehat{u}}({\bm{d}})$ 
 & $0.0278$                       & $0.0510$                       & $0.1166$              \\
\hline
 ${\alpha}{\big (}{\widehat{u}}({\bm{d}}){\big )}$ 
 & $3.2480(>K_{\rm{C}})$          & $3.0055(>K_{\rm{C}})$          & $2.7186(>K_{\rm{C}})$ \\
\hline
\hline
 ${\bm{d}}$ 
 & Fig.{\ref{Figure03}}(d)     & Fig.{\ref{Figure03}}(e)     & Fig.{\ref{Figure03}}(f)     \\
\hline
 ${\widehat{u}}({\bm{d}})$ 
 & $0.3371$                       & $0.1767$                       & $0.1949$              \\
\hline
 ${\alpha}{\big (}{\widehat{u}}({\bm{d}}){\big )}$ 
 & $2.5050(<K_{\rm{C}})$          & $2.6050(>K_{\rm{C}})$          & $2.5826(<K_{\rm{C}})$ \\
\hline
\end{tabular}
\end{center}
\label{Table01}
\end{table}

\vspace{5.0mm}

{\section{Comparison with Conventional Maximum Likelihood Framework}} \label{sec: Conventional ML}

In this section, we describe the conventional scheme 
for hyperparameter estimation in the maximum likelihood framework
and compare it with our proposed scheme. 
The conventional scheme estimates the hyperparameters 
by maximizing a marginal likelihood.
Marginal likelihoods are defined by regarding ``the probability of data 
when hyperparameters are given'' 
as a likelihood function of hyperparameters when data are given.
It is computed by marginalizing 
a joint probability of parameters and observed data 
with respect to parameters when hyperparameters are given 
and is expressed in terms of the partition functions 
of our assumed posterior and prior probabilities.
However, in our present problem for image segmentation, 
our prior probabilistic model 
is assumed to be the Potts model 
and often has the first order phase transition 
at a transition point. 
In such situation, we explain how hyperparameters are estimated 
in the conventional maximum likelihood framework with LBP's.

Instead of eq.(\ref{Prior}), 
the prior probability 
of a labeling ${\bm{a}}$ is assumed to be
\begin{eqnarray}
{\Pr}\{{\bm{A}}={\bm{a}}|K\}
={\frac{1}{{\cal{Y}}(K)}}
{\prod_{\{i,j\}{\in}{\cal{E}}}}
{\exp}{\Big (}{\frac{1}{2}}K
{\delta}_{a_{i},a_{j}}{\Big )}
\label{Prior-ML}
\end{eqnarray}
where ${\cal{Y}}(K)$ is a normalization constant 
and corresponds to the partition function 
of our prior probabilistic model. 
By substituting 
eqs.(\ref{Prior-ML}) and 
(\ref{DegradationProcess}) 
into the Bayes formula, 
we derive the posterior probability distribution
\begin{eqnarray}
{\Pr}\{{\bm{A}}={\bm{a}}|
{\bm{D}}={\bm{d}},K,{\bm{\Theta}}\}
={\frac{1}{{\cal{Y}}({\bm{d}},K,{\bm{\Theta}})}}
{\Big (}{\prod_{i{\in}{\cal{V}}}}g({\bm{d}}_{i}|a_{i},{\bm{\Theta}}){\Big )}
{\Big (}{\prod_{\{i,j\}{\in}{\cal{E}}}}
{\exp}{\big (}{\frac{1}{2}}{\cal{K}}{\delta}_{a_{i},a_{j}}{\big )}{\Big )},
\label{Posterior-ML}
\end{eqnarray}
where ${\cal{Y}}(K,{\bm{d}},{\bm{\Theta}})$ 
is a normalization constant 
and corresponds to the partition function 
of our posterior probabilistic model. 

In the conventional maximum likelihood frameworks, 
estimation of hyperparameters ${\widehat{K}}({\bm{d}})$ and 
${\widehat{\bm{\Theta}}}({\bm{d}})
\equiv \{{\widehat{\bm{m}}}({\xi},{\bm{d}}),
{\widehat{{\bm{C}}}}({\xi},{\bm{d}})|{\xi}{\in}{\cal{Q}}\}$, 
for $K$ and 
${\bm{\Theta}} \equiv \{{\bm{m}}({\xi}),{\bm{C}}({\xi})\}$ 
are determined by maximizing the marginal likelihood 
${\Pr}{\big \{}{\bm{D}}={\bm{d}}{\big |}K,{\bm{\Theta}}{\big \}}$ as follows:
\begin{eqnarray}
{\big (}{\widehat{K}}({\bm{d}}),{\widehat{\bm{\Theta}}}({\bm{d}}){\big )}
={\arg}{\max_{(K,{\bm{\Theta}})}}
{\Pr}{\big \{}{\bm{D}}={\bm{d}}{\big |}K,{\bm{\Theta}}{\big \}},
\label{MML}
\end{eqnarray}
where
\begin{eqnarray}
{\Pr}{\big \{}{\bm{D}}={\bm{d}}{\big |}K,{\bm{\Theta}}{\big \}}
\equiv
{\sum_{\bm{z}}}
{\Pr}{\big \{}{\bm{D}}={\bm{d}}{\big |}{\bm{A}}={\bm{z}},{\bm{\Theta}}{\big \}}
{\Pr}{\big \{}{\bm{A}}={\bm{z}}{\big |}K{\big \}}.
\label{MarginalLikelihood-MML}
\end{eqnarray}
Maximization of marginal likelihood in eq.(\ref{MML}) 
can be rewritten as 
\begin{eqnarray}
{\widehat{K}}({\bm{d}})
& =& {\arg}{\max_{K}}
{\Pr}{\big \{}{\bm{D}}={\bm{d}}{\big |}K,{\bm{\Theta}}(K,{\bm{d}}){\big \}},
\label{MML-KHat}
\\
{\widehat{\bm{\Theta}}}({\bm{d}}) & = & {\bm{\Theta}}({\bm{d}},{\widehat{K}}({\bm{d}})),
\label{MML-ThetaHat}
\end{eqnarray}
where the set of hyperparameters 
${\bm{\Theta}}(K,{\bm{d}})
\equiv
\{{\bm{m}}({\xi},K,{\bm{d}}),{\bm{C}}({\xi},K,{\bm{d}})
|{\xi}{\in}{\cal{Q}}\}$ is 
determined so as to satisfy the 
following simultaneous fixed point equations: 
\begin{eqnarray}
{\frac{
{\displaystyle{{\sum_{i{\in}{\cal{V}}}}}}{\bm{d}}_{i}
{\Pr}\{A_{i}={\xi}|
{\bm{D}}={\bm{d}},K,{\bm{\Theta}}(K,{\bm{d}})\}
}
{
{\displaystyle{{\sum_{i{\in}{\cal{V}}}}}}
{\Pr}\{A_{i}={\xi}|
{\bm{D}}={\bm{d}},K,{\bm{\Theta}}(K,{\bm{d}})\}
}}
={\bm{m}}({\xi},K,{\bm{d}})
~({\xi}{\in}{\cal{Q}}),
\label{Determination-m-posterior-ML}
\end{eqnarray}
\begin{eqnarray}
& &{\hspace{-1.0cm}}
{\frac{
{\displaystyle{{\sum_{i{\in}{\cal{V}}}}}}
{\big (}{\bm{d}}_{i}-{\bm{m}}({\xi},K,{\bm{d}}){\big )}
{\big (}{\bm{d}}_{i}-{\bm{m}}({\xi},K,{\bm{d}}){\big )}^{\rm{T}}
{\Pr}\{A_{i}={\xi}|
{\bm{D}}={\bm{d}},K,{\bm{\Theta}}(K,{\bm{d}})\}
}{
{\displaystyle{{\sum_{i{\in}{\cal{V}}}}}}
{\Pr}\{A_{i}={\xi}|
{\bm{D}}={\bm{d}},K,{\bm{\Theta}}(K,{\bm{d}})\}
}}
\nonumber\\
& &{\hspace{7.50cm}}
={\bm{C}}({\xi},K,{\bm{d}})
{\hspace{2.0mm}}({\xi}{\in}{\cal{Q}}),
\label{Determination-C-posterior-ML}
\end{eqnarray}
for various values of $K$.
Equations (\ref{Determination-m-posterior-ML})
and (\ref{Determination-C-posterior-ML}) are derived 
by considering the extremum conditions 
of ${\ln}{\Pr}{\big \{}{\bm{D}}={\bm{d}}{\big |}K,{\bm{\Theta}}{\big \}}$
with respect to ${\bm{m}}({\xi})$ and ${\bm{C}}({\xi})$.
For each value of $K(>0)$, 
we compute ${\bm{\Theta}}(K,{\bm{d}})$ 
by solving the simultaneous fixed point equations 
(\ref{Determination-m-posterior-ML}) and (\ref{Determination-C-posterior-ML}) 
by means of the iterative numerical method.
Then we determine the estimates ${\widehat{K}}$ 
so as to maximize 
${\Pr}{\big \{}{\bm{D}}={\bm{d}}{\big |}K,{\bm{\Theta}}(K,{\bm{d}}){\big \}}$ 
with respect to $K$.
The estimate ${\widehat{\bm{a}}}({\bm{d}})
={\big (}{\widehat{a}}_{1}({\bm{d}}),{\widehat{a}}_{2}({\bm{d}}),{\cdots}
{\widehat{a}}_{|{\cal{V}}|}({\bm{d}}){\big )}^{\rm{T}}$ is 
determined by maximizing 
the marginal posterior probability distribution 
for each pixel $i({\in}{\cal{V}})$ as follows:
\begin{eqnarray}
{\widehat{a}}_{i}({\bm{d}})
={\arg}{\max_{a_{i}{\in}{\cal{Q}}}}
{\Pr}\{A_{i}=a_{i}|{\bm{D}}={\bm{d}},K,{\bm{\Theta}}\}
~(i{\in}{\cal{V}}),
\label{MPM-MML}
\end{eqnarray}
\begin{eqnarray}
{\Pr}\{A_{i}=a_{i}|{\bm{D}}={\bm{d}},K,{\bm{\Theta}}\}
\equiv
{\sum_{\bm{z}}}{\delta}_{a_{i},z_{i}}
{\Pr}\{{\bm{A}}={\bm{z}}|
{\bm{D}}={\bm{d}},K,{\bm{\Theta}}\}
~(i{\in}{\cal{V}}).
\label{MarginalPosterior-MML}
\end{eqnarray}

The left-hand sides of 
eqs.(\ref{Determination-m-posterior-ML})
and (\ref{Determination-C-posterior-ML}), 
the marginal posterior probability distribution 
${\Pr}\{A_{i}=a_{i}|{\bm{D}}={\bm{d}},K,{\bm{\Theta}}\}$ 
in eq.(\ref{MarginalPosterior-MML}), and the marginal likelihood 
${\Pr}{\big \{}{\bm{D}}={\bm{d}}{\big |}K,{\bm{\Theta}}{\big \}}$ 
in eq.(\ref{MarginalLikelihood-MML}) 
can be approximately computed 
by using the LBP for each set $(K,{\bm{\Theta}})$.
In the case of $q=8$, Fig.{\ref{Figure08}} shows 
the logarithm of marginal likelihood per pixel, 
${\frac{1}{|{\cal{V}}|}}
{\ln}{\Pr}{\big \{}{\bm{D}}={\bm{d}}
{\big |}K,{\bm{\Theta}}(K,{\bm{d}}){\big \}}$, 
in eqs.(\ref{MML-KHat})-(\ref{MML-ThetaHat}) 
for the observed images ${\bm{d}}$ 
in Figs.{\ref{Figure03}}(c) and (d).
${\widehat{K}}({\bm{d}})$ for Fig.{\ref{Figure03}}(c) is equal 
to ${\alpha}({\widehat{u}})$ obtained 
by our proposed scheme in {\S}3
and is larger than the first order transition point $K_{\rm{C}}$ 
of the $8$-state Potts model in the LBP.
On the other hand, 
${\widehat{K}}({\bm{d}})$ for Fig.{\ref{Figure03}}(d) 
is larger than 
${\alpha}({\widehat{u}})$ obtained by our proposed scheme in {\S}3
and is equal to $K_{\rm{C}}$.
These are typical cases of estimates obtained by 
our proposed scheme 
and the conventional maximum likelihood framework. 
In Fig.{\ref{Figure09}},
${\frac{1}{|{\cal{E}}|}}
{\displaystyle{{\sum_{\{i,j\}{\in}{\cal{E}}}}}}
{\displaystyle{{\sum_{{\bm{a}}}}}}
(1-{\delta}_{a_{i},a_{j}}){\Pr}{\big \{}{\bm{A}}={\bm{a}}
{\big |}{\bm{D}}={\bm{d}},K,{\bm{\Theta}}(K,{\bm{d}}){\big \}}$ 
is also shown for each observed image ${\bm{d}}$ 
in Figs.{\ref{Figure03}}(c) and (d).
We see that 
the derivative of the logarithm 
of marginal likelihood with respect to $K$ 
is equal to zero at the point satisfying 
\begin{eqnarray}
& &{\hspace{-1.0cm}}
{\sum_{\{i,j\}{\in}{\cal{E}}}}
{\sum_{{\zeta}{\in}{\cal{Q}}}}
{\sum_{{\zeta}'{\in}{\cal{Q}}}}
{\big (}1-{\delta}_{{\zeta},{\zeta}'}{\big )}
{\Pr}\{A_{i}={\zeta},A_{j}={\zeta}'|
{\bm{D}}={\bm{d}},{\widehat{K}}({\bm{d}}),
{\widehat{\bm{\Theta}}}({\bm{d}})\}
\nonumber\\
& &
=
{\sum_{\{i,j\}{\in}{\cal{E}}}}
{\sum_{{\zeta}{\in}{\cal{Q}}}}
{\sum_{{\zeta}'{\in}{\cal{Q}}}}
{\big (}1-{\delta}_{{\zeta},{\zeta}'}{\big )}
{\Pr}\{A_{i}={\zeta},A_{j}={\zeta}'|
{\widehat{K}}({\bm{d}})\}.
\label{Determination-K-posterior-ML}
\label{MarginalLikelihood-ExtremumPointK}
\end{eqnarray}
and it corresponds to the intersection between 
the black and the red solid line in Fig.{\ref{Figure09}}(a). 
The intersection corresponds also to the estimate 
of $({\alpha}({\widehat{u}}),{\widehat{u}})$ based on 
eq.(\ref{Determination-u-posterior}) in our proposed scheme.
In Figs.{\ref{Figure08}}(a) and {\ref{Figure09}}(a), 
the derivative of the logarithm of marginal likelihood 
with respect to $K$ 
is equal to zero at the maximum point of the marginal likelihood; 
and it means that the estimate of hyperparameter 
in our proposed scheme 
based on eq.(\ref{Determination-u-posterior}) 
is equivalent to the one 
in the conventional maximum likelihood estimation.
However, in Fig.{\ref{Figure08}}(b), 
we see that the logarithm of marginal likelihood 
is not differentiable 
at ${\widehat{K}}({\bm{d}})=K_{\rm{C}}$ 
for the observed image ${\bm{d}}$ of Fig.{\ref{Figure03}}(d) 
and eq.(\ref{Determination-K-posterior-ML}) 
does not satisfies at this point, 
although the estimate of ${\alpha}({\widehat{u}})$ 
in our proposed scheme 
corresponds to the intersection between 
the black and the red solid line 
in Fig.{\ref{Figure09}}(b), 
This is one of the major differences 
between our proposed scheme 
and the conventional maximum likelihood framework. 
\begin{figure}
\begin{center}
\includegraphics[height=4.0cm, bb = 0 0 700 246]{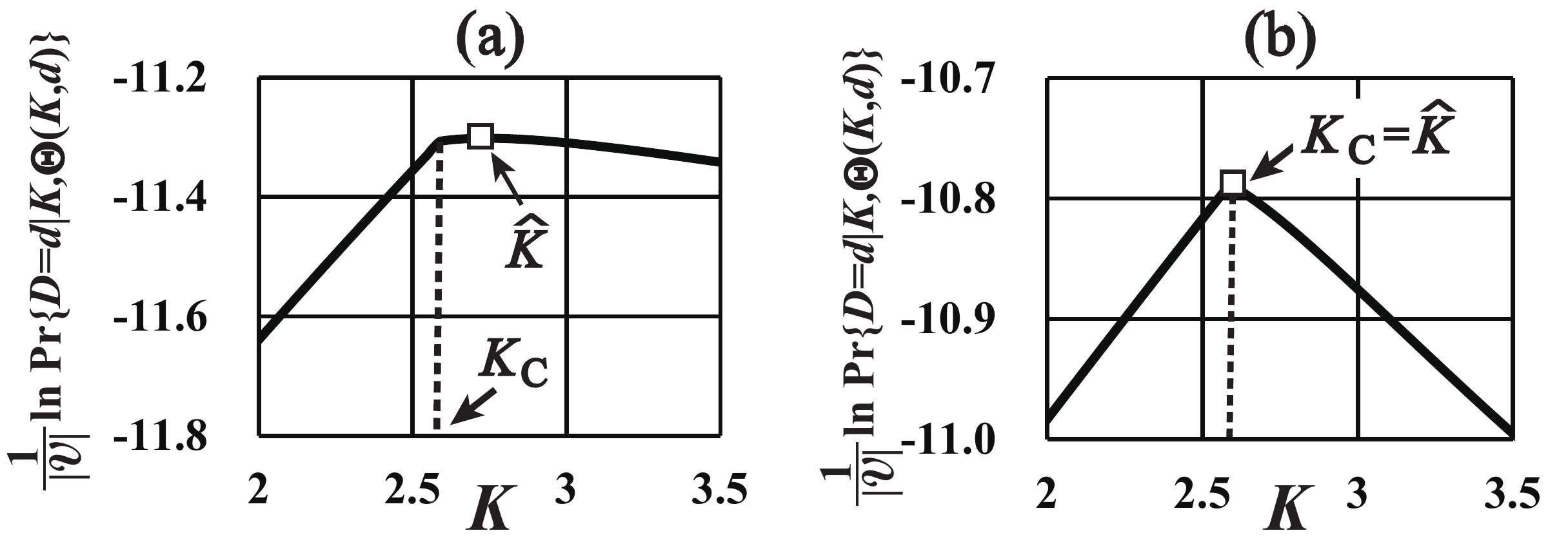}
\end{center}
\caption{
(a) and (b) are the logarithm of marginal likelihood 
${\frac{1}{|{\cal{V}}|}}
{\ln}{\Pr}{\big \{}{\bm{D}}={\bm{d}}
{\big |}K,{\bm{\Theta}}(K,{\bm{d}}){\big \}}$ 
for the observed images ${\bm{d}}$ 
in Fig.{\ref{Figure03}}(c) 
and Fig.{\ref{Figure03}}(d) 
in the case of $q=8$, which are shown as black solid curves.
$K_{\rm{C}}$ 
is the first order transition point 
of the $q$-state Potts model 
in the loopy belief propagation and 
is $2.5871$ for $q=8$, respectively.
}
\label{Figure08}
\end{figure}
\begin{figure}
\begin{center}
\includegraphics[height=4.0cm, bb = 0 0 689 332]{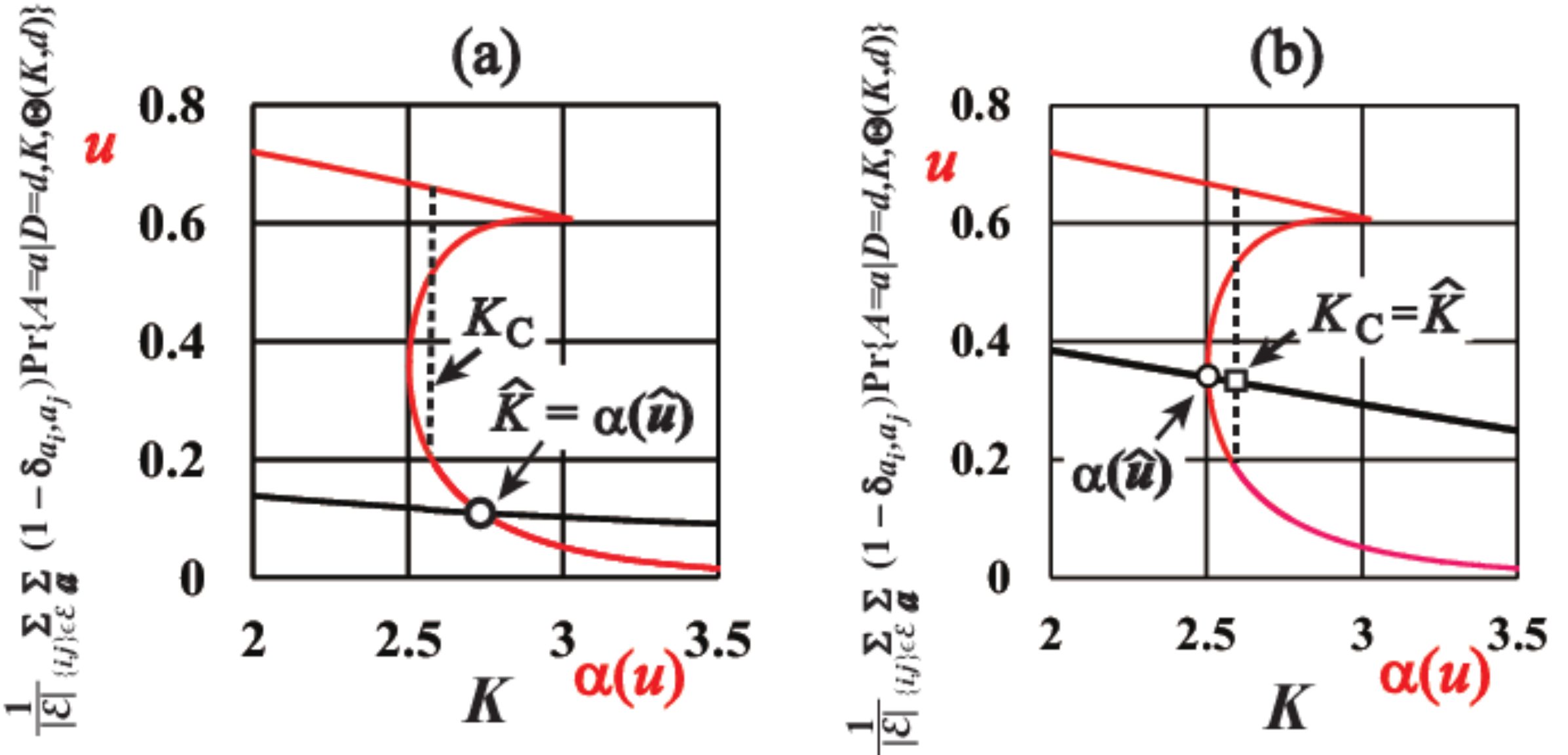}
\end{center}
\caption{
(a) and (b) are ${\frac{1}{|{\cal{E}}|}}
{\displaystyle{{\sum_{\{i,j\}{\in}{\cal{E}}}}}}
{\displaystyle{{\sum_{{\bm{a}}}}}}
(1-{\delta}_{a_{i},a_{j}}){\Pr}{\big \{}{\bm{A}}={\bm{a}}
{\big |}{\bm{D}}={\bm{d}},K,{\bm{\Theta}}(K,{\bm{d}}){\big \}}$ 
for the observed images ${\bm{d}}$ 
in Fig.{\ref{Figure03}}(c) 
and Fig.{\ref{Figure03}}(d) 
in the case of $q=8$, which are shown as black solid curves.
The red solid curves are $({\alpha}(u),u)$ 
by using the loopy belief propagation 
in eqs.(\ref{Prior}) and (\ref{PriorConstraint}). 
$K_{\rm{C}}$ 
is the first order transition point 
of the $q$-state Potts model 
in the loopy belief propagation and 
is $2.5871$ for $q=8$, respectively.
}
\label{Figure09}
\end{figure}

\vspace{5.0mm}

{\section{Concluding Remarks}} \label{sec: ConcludingRemarks}

In the present paper,
we proposed 
a Bayesian image segmentation model 
based on Potts prior. Under the segmentation model, 
we then proposed a hyperparameter estimation scheme 
based on conditional maximization 
for entropy of the prior, 
and gave the practical inference algorithm 
based on LBP.

The conventional maximum likelihood framework, 
which is based on the maximization of marginal likelihood, 
is constructed 
from the free energies 
of the prior and the posterior probabilities.
In the present paper, 
the prior probability is assumed to be the Potts model 
and it has the first order phase transition 
on computing some statistical quantities by means of the LBP.
Because the derivative of free energy has discontinuity 
in the first order phase transition point, 
it is very difficult 
to search the maximum point via the extremum condition 
of the marginal likelihood 
with respect to some of the hyperparameters. 
Actually, ${\ln}{\Pr}\{{\bm{D}}={\bm{d}}|u,{\cal{G}}\}$ 
is given in terms of the normalization constants 
${\cal{Y}}({\bm{d}},K,{\bm{\Theta}})$ and ${\cal{Y}}(K)$ 
in eqs.(\ref{Prior-ML}) and (\ref{Posterior-ML}) 
as follows:
\begin{eqnarray}
{\ln}{\big (}
{\Pr}\{{\bm{D}}={\bm{d}}|
K,{\bm{\Theta}}\}{\big )}
=
{\ln}{\big (}{\cal{Y}}({\bm{d}},K,{\bm{\Theta}}){\big )}
-{\ln}{\big (}{\cal{Y}}(K){\big )}.
\label{MarginalLikelihoodRepresentation}
\end{eqnarray}
The logarithms 
$-{\ln}{\cal{Y}}(K)$ 
and 
$-{\ln}{\cal{Y}}({\bm{d}},K,{\bm{\Theta}})$ 
correspond to the free energies 
of the posterior and the prior probabilistic models 
in eqs.(\ref{Prior-ML}) and (\ref{Posterior-ML}), respectively. 
As shown in Fig.{\ref{Figure01}},
the LBP of the $q$-state Potts prior (\ref{Prior-ML}) for $q{\ge}3$ 
with no external fields have the first order phase transition. 
In addition, 
the free energy 
$f(K)=-{\frac{1}{|{\cal{V}}|}}{\ln}{\cal{Y}}(K)$ per pixel 
has at least one singular point $K=K_{\rm{C}}$ 
at which the derivative ${\frac{d}{dK}}f(K)$ is discontinuous 
with respect to $K$.
Although one of the useful procedures 
for realizing the maximization 
of marginal likelihood 
is the EM algorithm{\cite{DempsterLairdRubinRoyal1977}}, 
it is based on the analysis for hyperparameters 
and is hard to be adopted 
in the conventional maximum likelihood framework. 

Our proposed algorithm in {\S}3 
is based on the constrained maximization 
of the entropies 
in eqs.(\ref{Prior-EntropyMax}) 
and (\ref{Posterior-EntropyMax})
without using the maximization 
of marginal likelihood 
${\Pr}{\big \{}{\bm{D}}={\bm{d}}{\big |}
K,{\bm{\Theta}}{\big \}}$ 
in eqs.(\ref{MarginalLikelihood-MML}) 
and (\ref{MarginalLikelihoodRepresentation}).
Particularly, 
with the $q$-state Potts model 
in eqs.(\ref{Prior}) and (\ref{PriorConstraint}), 
the interaction parameter ${\alpha}(u)$ 
of the $q$-state Potts model (\ref{Prior}) 
is a one-valued function of $u$ 
which corresponds to the internal energy 
$-{\frac{1}{2}}
{\displaystyle{{\sum_{\{i,j\}{\in}{\cal{E}}}}}}
{\displaystyle{{\sum_{\bm{z}}}}}{\delta}_{z_{i},z_{j}}
{\Pr}\{{\bm{A}}={\bm{z}}|u\}$, 
when ${\alpha}(u)$ is regarded 
as the inverse temperature of the system.
It is the key to the success 
of our iterative inference algorithm (in {\S}3) 
on estimating
the average vectors ${\bm{m}}({\xi})$, 
covariance matrices ${\bm{C}}({\xi})$ (${\xi}{\in}{\cal{Q}}$),  
$u$, and ${\alpha}(u)$
in eqs.(\ref{Posterior})-(\ref{Posterior-PartitionFunction}),
as shown in Figs.{\ref{Figure04}} and {\ref{Figure05}} 
and Table {\ref{Table01}}.

In {\S}4, we have conducted the maximization 
of marginal likelihood 
${\Pr}{\big \{}{\bm{D}}={\bm{d}}{\big |}
K,{\bm{\Theta}}{\big \}}$ 
in eqs.(\ref{MarginalLikelihood-MML}) 
and (\ref{MarginalLikelihoodRepresentation}) and compare it 
with our proposed algorithm. 
The extremum conditions for 
average vectors ${\bm{m}}({\xi})$ 
and 
covariance matrices ${\bm{C}}({\xi})$ (${\xi}{\in}{\cal{Q}}$) 
have been given by 
eqs.
(\ref{Determination-m-posterior-ML})
and (\ref{Determination-C-posterior-ML}). 
They are basically equivalent to 
the constraints 
(\ref{Determination-m-posterior})
and (\ref{Determination-C-posterior}) 
in our constrained maximization of entropies 
in eqs.(\ref{Prior-EntropyMax}) 
and (\ref{Posterior-EntropyMax}) in {\S}2 and {\S}3. 
However, their difference is in eq.(\ref{MML-KHat}).
As mentioned above, 
${\Pr}{\big \{}{\bm{D}}={\bm{d}}{\big |}
K,{\bm{\Theta}}{\big \}}$ is not differentiable 
at $K=K_{\rm{C}}$, and therefore 
the extremum condition of 
${\Pr}{\big \{}{\bm{D}}={\bm{d}}{\big |}
K,{\bm{\Theta}}{\big \}}$ 
with respect to $K$ cannot be considered as its maximization
when ${\widehat{K}}({\bm{d}})$ is equal to $K_{\rm{C}}$. 
On the other hand, 
if ${\widehat{K}}({\bm{d}})$ is equal to $K_{\rm{C}}$, 
we can consider 
the extremum condition of 
${\Pr}{\big \{}{\bm{D}}={\bm{d}}{\big |}
K,{\bm{\Theta}}{\big \}}$ 
and reduce the deterministic equation 
of ${\widehat{K}}({\bm{d}})$ to 
eq.(\ref{Determination-K-posterior-ML}).
Equation (\ref{Determination-K-posterior-ML}) 
is equivalent to 
eq.(\ref{Determination-u-posterior}).
In this case, 
the conventional maximum likelihood framework 
in {\S}4 
is equivalent to 
the constrained maximum entropy framework 
in {\S}2 and {\S}3.
The segmentation result for Fig.{\ref{Figure03}}(c) 
in the case of $q=8$ is one of the typical examples,  
where we obtain 
${\alpha}{\big (}{\widehat{u}}({\bm{d}}){\big )}
={\widehat{K}}({\bm{d}}) > K_{\rm{C}}$; 
and the estimates ${\bm{\widehat{\Theta}}}({\bm{d}})$ 
by using our proposed algorithm 
are equal to each other, 
as shown in Table {\ref{Table01}}.

However, in order to know if ${\widehat{K}}({\bm{d}})$ 
is equal to $K_{\rm{C}}$ 
in the conventional maximum likelihood framework, 
we have to compute 
${\Pr}{\big \{}{\bm{D}}={\bm{d}}{\big |}
K,{\bm{\Theta}}(K,{\bm{d}}){\big \}}$ 
to satisfy the simultaneous fixed point equations 
(\ref{Determination-m-posterior-ML})
and (\ref{Determination-C-posterior-ML})
with respect to ${\bm{\Theta}}(K,{\bm{d}})$ 
for various values of $K$, 
as shown in Fig.{\ref{Figure08}}(b). 
This is the main difficulty for 
achieving the conventional maximization of marginal likelihood 
${\Pr}{\big \{}{\bm{D}}={\bm{d}}{\big |}
K,{\bm{\Theta}}{\big \}}$ 
in eqs.(\ref{MarginalLikelihood-MML}) 
and (\ref{MarginalLikelihoodRepresentation}), 
although our proposed algorithm 
is constructed from just iterative procedures 
with respect to ${\widehat{u}}({\bm{d}})$, 
${\alpha}{\big (}{\widehat{u}}({\bm{d}}){\big )}$ 
and ${\bm{\widehat{\Theta}}}({\bm{d}})$,
as shown in {\it ``Inference Algorithm 
for ${\widehat{u}}({\bm{d}})$, 
${\alpha}{\big (}{\widehat{u}}({\bm{d}}){\big )}$ 
and ${\widehat{\bm{\Theta}}}({\bm{d}})$''} of {\S}3.

Finally, we discuss the relationship between 
the proposed framework and the graph cut method. 
One may consider using a graph cut method to achieve 
the image segmentations by means of the MRF.
The graph cut methods can derive  
the exact global maximum configuration ${\bm{a}}^{\bm{*}}$ 
of the posterior probabilistic distribution 
${\Pr}\{{\bm{A}}={\bm{z}}|{\bm{D}}={\bm{d}},K,{\bm{\Theta}}\}$:
\begin{eqnarray}
{\bm{a}}^{\bm{*}}
={\arg}{\max_{{\bm{z}}}}
{\Pr}\{{\bm{A}}={\bm{z}}|
{\bm{D}}={\bm{d}},K,{\bm{\Theta}}\},
\label{MAP-ML}
\end{eqnarray}
at least for $q=2${\cite{BoykovVekslerZabih2001}}, 
and recently it has been extended 
to an approximate graph cut method which can be applied also to 
the case of $q{\ge}3${\cite{AlahariKohliTorr2011}}.
However, the graph cut method cannot give the estimates of 
hyperparameters $K$ and ${\bm{\Theta}}$ 
from one single observed image ${\bm{d}}$.
Instead, the hyperparameter $K$ of the Potts prior 
is usually estimated 
by using supervised learning from many labeled pairs 
$({\bm{a}}^{(1)},{\bm{d}}^{(1)}),
({\bm{a}}^{(2)},{\bm{d}}^{(2)}),{\cdots},
({\bm{a}}^{(N)},{\bm{d}}^{(N)})$, 
where the labeled image ${\bm{a}}^{(n)}$ is the ground truth 
for each observed image ${\bm{d}}^{(n)}$ for $n=1,2,{\cdots},N$.
When the supervised learning approaches are included
in the graph cut method for image segmentation, 
the following maximum likelihood estimation 
is often used for hyperparameter estimation: 
\begin{eqnarray}
(K^{*},{\bm{\Theta}}^{\bm{*}})
= 
{\arg}{\max_{(K,{\bm{\Theta}})}}
{\sum_{n=1}^{N}}
{\ln}{\big (}{\Pr}\{{\bm{A}}={\bm{a}}^{(n)},
{\bm{D}}={\bm{d}}^{(n)}|K,{\bm{\Theta}}\}{\big )}.
\label{MAP-ML-Learning}
\end{eqnarray}
To sum up, we have clarified the theoretical relationship 
between the LBP and the graph cut method 
and have proposed novel statistical methods 
for probabilistic image segmentations by means of the MRF.

\vspace{5.0mm}

\section*{Acknowledgements}
This work was partly supported 
by the Grants-In-Aid (No.25280089, No.25120009 and No.24700220) 
for Scientific Research from the Ministry of Education, 
Culture, Sports, Science and Technology of Japan.

\end{document}